\documentclass[letterpaper,10pt,twocolumn]{IEEEtran}
\usepackage[mathscr]{eucal}
\usepackage[cmex10]{amsmath}
\usepackage{epsfig,epsf,psfrag,amssymb,amsfonts,latexsym,url,amsmath,graphicx,bm,cite, xcolor,url}
\usepackage{fixltx2e}
\usepackage{verbatim}
\usepackage{bm}
\usepackage[footnotesize]{subfigure}
\usepackage{latexsym, amssymb, amsmath}

\DeclareMathOperator\Deg{deg}
\newcommand{\bd}{\begin{definition}}
\newcommand{\ed}{\end{definition}}
\def\bcase{\vskip .1cm \begin{numcases}}
\def\ecase{\end{numcases} \vskip .1cm \noindent}

\def\0{{\mathbf{0}}}
\def\beq{\begin{equation}}
\def\eeq{\end{equation}}
\def\beqn{\vskip .1cm \begin{eqnarray}}
\def\eeqn{\end{eqnarray} \vskip .1cm \noindent}
\def\beqnn{\vskip .01cm \begin{eqnarray*}}
\def\eeqnn{\end{eqnarray*} \vskip .01cm \noindent}

\def\nn{\nonumber}

\newcommand{\calO}{\mathcal{O}}

\newcommand{\rhobf}{\bm{\rho}}

\def\calC{{\mathcal{C}}}
\newcommand{\calH}{\mathcal{H}}
\newcommand{\calL}{\mathcal{L}}

\newcommand{\tilJ}{\widetilde{J}}

\newcommand{\tilK}{\widetilde{K}}
\newcommand{\calS}{\mathcal{S}}
\newcommand{\calN}{\mathcal{N}}
\newcommand{\calE}{\mathcal{E}}

\newcommand{\calV}{\mathcal{V}}
\newcommand{\calB}{\mathcal{B}}
\newcommand{\calT}{\mathcal{T}}
\newcommand{\calA}{\mathcal{A}}

\newcommand{\bx}{\mathbf{x}}
\newcommand{\bR}{\mathbb{R}}
\newcommand{\bN}{\mathbb{N}}

\newcommand{\nbd}{\mathrm{nbd}}
\newcommand{\calP}{\mathcal{P}}

\newcommand{\bM}{\mathbf{M}}
\newcommand{\bP}{\mathbb{P}}
\newcommand{\bE}{\mathbb{E}}

\newcommand{\bpi}{\bm{\pi}}
\newcommand{\bSigma}{\bm{\Sigma}}
\newcommand{\bsigma}{\bm{\sigma}}
\newcommand{\bDelta}{\bm{\Delta}}
\newcommand{\bmeta}{\bm{\eta}}

\newcommand{\bC}{\mathbf{C}}
\newcommand{\bzero}{\mathbf{0}}

\newcommand{\hp}{\widehat{p}}
\newcommand{\argmin}{\operatornamewithlimits{argmin}}
\newcommand{\argmax}{\operatornamewithlimits{argmax}}

\newcommand{\hrho}{\widehat{\rho}}

\newcommand{\Tr}{\mathrm{Tr}}

\newcommand{\Path}{\mathrm{Path}}

\newcommand{\var}{\mathrm{Var}}
\newcommand{\diam}{\mathrm{diam}}

\newcommand{\hcalE}{\calE_{{\text{\tiny CL}}}}
\newcommand{\ie}{{\it i.e.}}

\newtheorem{theorem}{Theorem}
\newtheorem{lemma}[theorem]{Lemma}
\newtheorem{proposition}[theorem]{Proposition}
\newtheorem{corollary}[theorem]{Corollary}

\newtheorem{definition}{Definition}

\newenvironment{myproof}{\noindent{\em Proof:} \hspace*{1em}}{
    \hspace*{\fill} $\Box$ }

\newcommand{\bprf}{\begin{myproof}}
\newcommand{\eprf}{\end{myproof}}

\author{Vincent Y.~F.~Tan,~\IEEEmembership{Student Member,~IEEE,}
        Animashree Anandkumar,~\IEEEmembership{Member,~IEEE,} \\ Alan S. Willsky,~\IEEEmembership{Fellow,~IEEE.}
          \thanks{\scriptsize The authors are with Department of Electrical Engineering and Computer Science and  the
        Stochastic Systems Group, Laboratory for Information and Decision Systems,
        Massachusetts Institute of Technology. Email: {\tt\{vtan, animakum, willsky\}@mit.edu}. }
 \thanks{\scriptsize This work is supported in part by a AFOSR through Grant FA9550-08-1-1080, in part by a MURI funded through
ARO Grant W911NF-06-1-0076  and in part under a MURI through AFOSR Grant FA9550-06-1-0324. Vincent Tan is also supported by A*STAR, Singapore. }
 \thanks{\scriptsize This work was presented in part at the Allerton Conference on Communication, Control, and Computing, Monticello, IL, Sep 2009. }}

\title{Learning Gaussian Tree Models: Analysis of Error Exponents and Extremal Structures}

\begin{document}
\maketitle

\begin{abstract}
The problem of learning tree-structured Gaussian
graphical models from independent and identically distributed (i.i.d.) samples is considered. The influence of
the tree structure and the parameters of the Gaussian distribution
on the learning rate as the number of samples increases is discussed. Specifically, the error exponent
corresponding to the event that the estimated tree structure differs
from the actual unknown tree structure of the   distribution is
analyzed. Finding the error exponent reduces to a least-squares problem
in the very noisy learning regime. In this regime, it is shown
that the extremal tree structure that  minimizes the  error exponent is the star for any fixed set of correlation coefficients on the edges of
the tree.  If the magnitudes of all the correlation coefficients are less than 0.63, it is also shown that the tree structure that maximizes the error exponent is the Markov chain.   In other words, the star and the chain graphs represent the
hardest and the easiest structures to learn in the class of tree-structured Gaussian
graphical models. This result can also be intuitively explained by
correlation decay: pairs of nodes which are far apart, in terms of
graph distance, are unlikely to be mistaken as edges by the
maximum-likelihood estimator in the asymptotic regime.
\end{abstract}



\IEEEpeerreviewmaketitle

\section{Introduction}
Learning of structure and interdependencies of a large collection of
random variables from a set of data samples is an important
  task in signal and image analysis and many other scientific domains (see examples
in~\cite{Pearl, Gei94, Lau96, Wai08} and references therein). This task is
extremely challenging when the dimensionality of the data is large
compared to the number of samples. Furthermore, structure learning
of multivariate distributions is also complicated as it is
imperative to find the right balance between data fidelity and
overfitting the data to the model. This problem is circumvented when
we limit the  distributions to the set of Markov tree
distributions,   which have a fixed number of parameters and are
  tractable for learning~\cite{CL68} and statistical
inference~\cite{Pearl,Wai08}.

The problem of maximum-likelihood (ML) learning of a Markov tree
distribution from i.i.d.\ samples has an elegant  solution,
 proposed by Chow and Liu in~\cite{CL68}. The ML
tree structure is given by the maximum-weight spanning tree (MWST)
with empirical mutual information quantities as the edge weights.
Furthermore, the ML algorithm is {\em consistent}~\cite{Cho73}, which implies that the error probability in
learning the tree structure decays to zero with the number of samples
available for learning.

While consistency is an important qualitative property, there is substantial motivation for additional and more quantitative characterization of performance. One such measure, which we investigate in this theoretical paper is the     rate of decay of the error probability, \ie, the probability that
the ML  estimate of the edge set differs from the true edge set. When the error probability decays exponentially, the learning rate is usually referred to as the {\em
error exponent}, which provides a careful measure of performance
of the learning algorithm since a larger rate implies a faster decay
of the error probability.


We answer three fundamental questions in this paper.  (i) Can we
characterize the error exponent for structure learning by the ML
algorithm for  tree-structured Gaussian graphical
models (also called Gauss-Markov random fields)? (ii) How do the {\em structure}
and {\em parameters} of the model influence the error exponent?
(iii) What are extremal tree distributions for learning, \ie, the distributions
that  maximize and minimize the error exponents?   We believe
that our intuitively appealing answers to these important questions
provide key insights for learning tree-structured Gaussian graphical
models from data, and thus, for modeling high-dimensional data using
parameterized tree-structured distributions.


\subsection{Summary of Main Results}

We derive the error exponent as the optimal  value of the objective function of a non-convex
optimization problem, which can only be solved numerically (Theorem~\ref{thm:dom}). To gain
better insights into when errors occur, we approximate the error exponent with a closed-form
expression that can be interpreted as the signal-to-noise ratio
(SNR) for structure learning (Theorem~\ref{thm:euc_gauss}), thus
showing  how the parameters of the true
model affect learning. Furthermore, we show that   due to {\em
correlation decay}, pairs of nodes which are far apart, in terms of
their graph distance, are unlikely to be mistaken as  edges by the ML
estimator. This is not only an intuitive result, but also results
in a significant reduction in the computational complexity to find the
exponent --  from $\calO(d^{d-2})$  for  exhaustive search and $\calO(d^3)$ for discrete tree models~\cite{Tan09} to $\calO(d)$ for Gaussians (Proposition~\ref{lem:facttilK}), where $d$ is the number of nodes.

We then analyze  extremal tree structures for learning, given a fixed
set of correlation coefficients on the edges of the tree. Our main
result is the following: The {\em star}  graph
 minimizes the error exponent and if the absolute value of all the correlation coefficients of the variables along the edges is less than 0.63, then the {\it Markov chain} also maximizes the error exponent 
(Theorem~\ref{thm:graph_topo}). Therefore, the extremal tree structures
in terms of the diameter are {\em also}  extremal trees for learning
Gaussian tree distributions.   This agrees with the intuition that  the amount of
correlation decay increases with the tree diameter, and that
correlation decay helps the ML estimator to better distinguish the
edges from the non-neighbor pairs. Lastly, we analyze how changing the size of the tree influences the magnitude of the error exponent (Propositions~\ref{prop:subtree} and \ref{prop:supertree}).

\subsection{Related Work}
There is a substantial body of work on approximate learning of graphical models (also known as Markov random fields) from data {\it e.g.}~\cite{Chi02, Dud04, Mei06, Wai06}. The authors of these papers use various score-based approaches~\cite{Chi02}, the maximum entropy principle~\cite{Dud04} or $\ell_1$ regularization~\cite{Mei06, Wai06} as approximate structure learning techniques. Consistency guarantees in terms of the number of samples, the number of variables and the maximum neighborhood size are provided. Information-theoretic limits~\cite{San08} for learning graphical models have also been derived. In~\cite{Zuk06}, bounds on the error rate for learning the structure of Bayesian networks were provided but in contrast to our work, these bounds are not asymptotically tight (cf.\ Theorem~\ref{thm:dom}). Furthermore, the analysis in~\cite{Zuk06} is tied to the Bayesian Information Criterion. The focus of our paper is the analysis of the Chow-Liu~\cite{CL68} algorithm as an {\em exact} learning technique for estimating the tree structure and comparing error rates amongst different graphical models. In a recent paper~\cite{Mon09}, the authors concluded that if the graphical model possesses long range correlations, then it is difficult to learn. In this paper, we in fact identify the extremal structures and distributions in terms of error exponents for structure learning.  The area of study in statistics known as {\em covariance selection}~\cite{Dem72,dAsp08} also has connections with structure learning in Gaussian graphical models. Covariance selection involves estimating the non-zero elements in the inverse covariance matrix and providing consistency guarantees of the estimate in some norm, {\it e.g.} the Frobenius norm in~\cite{Roth08}.  

We previously analyzed the error exponent for learning discrete tree distributions in~\cite{Tan09}. We proved that for every discrete spanning tree model, the error
exponent for learning is strictly positive, which implies that the
error probability decays exponentially fast. In this paper, we
extend these results to Gaussian tree models and derive
new results which are both explicit and intuitive by exploiting the
properties of  Gaussians. The results we obtain in
Sections~\ref{sec:ee} and~\ref{sec:euc_gauss} are analogous to the
results in~\cite{Tan09} obtained for discrete
distributions, although the proof techniques are  different. Sections~\ref{sec:simplifyKp} and~\ref{sec:structure} contain  new results thanks to simplifications which hold  for Gaussians but which do not  hold for discrete distributions.

\subsection{Paper Outline}
This paper is organized as follows: In Section~\ref{sec:prelims}, we
state the problem precisely and provide  necessary  preliminaries
on learning Gaussian tree  models. In Section~\ref{sec:ee}, we derive an expression for the so-called crossover rate of two pairs of nodes. We then relate the set of
crossover rates to the error exponent for learning the tree structure. In Section~\ref{sec:euc_gauss}, we leverage on ideas from Euclidean information theory~\cite{Bor08} to state conditions
that allow accurate approximations of the error exponent. We demonstrate in
Section~\ref{sec:simplifyKp} how  to
 reduce the computational complexity for calculating the exponent.
 In Section~\ref{sec:structure}, we identify extremal structures that
 maximize and minimize the error exponent. Numerical results are presented in Section~\ref{sec:num} and we conclude the discussion in Section~\ref{sec:concl}. 

\section{Preliminaries and Problem Statement} \label{sec:prelims}
\subsection{Basics of Undirected Gaussian Graphical Models} \label{sec:graph_models}
{\em Undirected graphical models} or {\em Markov random fields}\footnote{In this paper, we use the terms ``graphical models'' and ``Markov random fields'' interchangeably.}  (MRFs) are probability distributions that
factorize according to given undirected graphs~\cite{Lau96}. In this
paper, we focus solely on {\em spanning trees} (\ie, undirected, acyclic,
connected graphs). A $d$-dimensional random vector $\bx=[
x_1,\ldots, x_d  ]^T \in \bR^d$ is said to be {\em Markov} on a
spanning tree  $T_p=(\calV,\calE_p)$ with   vertex (or node)
set $\calV=\{1,\ldots, d\}$ and  edge  set
$\calE_p\subset \binom{\calV}{2}$ if its distribution  $p(\bx)$
satisfies the (local) Markov property: $ p(x_i|x_{\calV \setminus \{i\}}) = p(x_i|x_{\nbd(i)}),$
where $\nbd(i):=\{j\in\calV :(i,j)\in \calE_p\}$ denotes the set of neighbors of node $i$.  We also denote the set of spanning trees with $d$ nodes as $\calT^d$, thus $T_p\in \calT^d$. Since $p$ is Markov on the tree $T_p$, its  probability density function (pdf) factorizes according to $T_p$ into node marginals $\{p_i:i\in\calV\}$ and pairwise marginals $\{p_{i,j}:(i,j)\in\calE_p\}$ in the following specific  way~\cite{Lau96} given the edge set  $\calE_p$:
\begin{equation}
p(\bx)~=~\prod_{i\in\calV}p_i(x_i)\prod_{(i,j)\in\calE_p}\frac{p_{i,j}(x_i,x_j)}{p_{i}(x_i) p_{j}(x_j)},
\label{eqn:tree_decomp}
\end{equation}
We assume that $p$, in addition to being Markov on the
spanning tree $T_p=(\calV,\calE_p)$, is a {\em Gaussian graphical
model} or {\em Gauss-Markov random field} (GMRF) with    known zero mean\footnote{Our results also extend to the scenario
where the mean of the Gaussian is unknown and has
to be estimated from the samples.}  and   unknown
positive definite covariance matrix $\bSigma\succ 0$. Thus, $p(\bx)$
can be written as
\begin{equation}
p(\bx)~=~\frac{1}{ (2\pi)^{d/2} |\bSigma|^{1/2}}  \exp \left(- \frac{1}{2} \bx^T \bSigma^{-1} \bx \right).\label{eqn:gaussian_form}
\end{equation}
We  also use the notation $p(\bx)=\calN(\bx;\0,\bSigma)$ as a
shorthand for~\eqref{eqn:gaussian_form}. For Gaussian graphical
models, it is known that the fill-pattern of the inverse covariance
matrix  $\bSigma^{-1}$ encodes
the structure of $p(\bx)$~\cite{Lau96}, \ie, $ \bSigma^{-1}(i,j) = 0
$ if and only if (iff) $(i,j) \notin \calE_p$.

We  denote the set of pdfs on $\bR^d$ by $\calP(\bR^d)$, the set of
Gaussian pdfs on $\bR^d$ by $\calP_{\calN}(\bR^d)$ and the
set of Gaussian graphical models which factorize according to some
tree in $\calT^d$ as $\calP_{\calN}(\bR^d, \calT^d)$.
For learning the structure  of $p(\bx)$ (or equivalently the fill-pattern of $\bSigma^{-1}$), we are provided with  a set of  $d$-dimensional samples $\bx^n :=\{ \bx_1, \ldots, \bx_n\}$  drawn  from $p$, where  $\bx_k:=[ x_{k,1}, \ldots,  x_{k,d}]^T \in \bR^d$. 

\subsection{ML Estimation of Gaussian Tree Models} \label{sec:chowliu}
In this subsection, we review the  Chow-Liu  ML learning
algorithm~\cite{CL68} for estimating the structure of $p$
given samples $\bx^n$. Denoting
$D(p_1||p_2):=\bE_{p_1} \log (p_1/p_2 ) $ as the
Kullback-Leibler (KL) divergence~\cite{Cov06} between $p_1$ and $p_2$, the
ML estimate of the  structure     $\hcalE(\bx^n)$ is given by the optimization
problem\footnote{Note that it is unnecessary to impose the
Gaussianity constraint on $q$
in~\eqref{eqn:cloptgauss}. We can optimize over $ \calP(\bR^d,
\calT^d)$ instead of  $\calP_{\calN}(\bR^d, \calT^d)$. It can be
shown that the optimal distribution is still  Gaussian. We omit the
proof for brevity.}
\begin{equation}
\hcalE(\bx^n) ~:=~ \argmin_{\calE_q : q\in \calP_{\calN}(\bR^d, \calT^d)}\,\, D(\hp\, ||\,  q), \label{eqn:cloptgauss}
\end{equation}
where $\hp(\bx):=\calN(\bx; \bzero, \widehat{\bSigma})$ and $\widehat{\bSigma}:=1/n\sum_{k=1}^n \bx_k \bx_k^T$
is the {\em empirical covariance matrix}.  Given  $\hp$, and exploiting the fact that $q$ in \eqref{eqn:cloptgauss} factorizes according to a tree as in~\eqref{eqn:tree_decomp}, Chow and Liu~\cite{CL68} showed that the optimization for the optimal edge set in~\eqref{eqn:cloptgauss} can be reduced to a MWST problem:
\begin{equation}
\hcalE(\bx^n)  ~=~ \argmax_{\calE_q : q\in \calP_{\calN}(\bR^d, \calT^d)} \,\,\sum_{e\in \calE_q} I(\hp_e), \label{eqn:mwst}
\end{equation}
where the edge weights are the {\em empirical mutual information
quantities}~\cite{Cov06}  given by\footnote{Our notation for the mutual information between two random variables
differs from the conventional one in~\cite{Cov06}.}
\begin{align}
I(\hp_e) = -\frac{1}{2}\log\left(1-\widehat{\rho}_e^2\right),  \label{eqn:simple_mi}
\end{align}
and where the {\em empirical correlation coefficients} are given by
$
\widehat{\rho}_{e}=\widehat{\rho}_{i,j} := {\widehat{\bSigma}(i,j)}/{ (\widehat{\bSigma}(i,i) \widehat{\bSigma}(j,j))^{1/2}}.
$
Note that in~\eqref{eqn:mwst}, the estimated edge set $\hcalE(\bx^n)$ depends on $n$ and, specifically, on the samples in $\bx^n$ and we make this dependence explicit. We assume that $T_p$ is a spanning tree because  with probability 1, the resulting optimization problem in \eqref{eqn:mwst} produces a spanning tree  as all the mutual information quantities in \eqref{eqn:simple_mi} will be non-zero. If $T_p$ {\em were} allowed to be a {\em proper forest} (a tree that is not connected), the estimation of $\calE_p $ will be inconsistent because the learned edge set will  be different from the true edge set.

\subsection{Problem Statement}
We now state our problem formally. Given a set of i.i.d.\ samples $\bx^n$ drawn from an unknown  Gaussian tree model $p$ with edge set $\calE_p$, we define  the error event that the set of edges is estimated incorrectly as
\begin{equation}
\calA_n~:=~\{\bx^n :\hcalE(\bx^n)  \ne \calE_p\}, \label{eqn:err}
\end{equation}
where $\hcalE(\bx^n)$ is the edge set of the Chow-Liu ML estimator in~\eqref{eqn:cloptgauss}.  In this paper, we are interested to {\em compute} and subsequently {\em study} the  {\em error exponent} $K_p$, or the rate that the error probability of the event $\calA_n$ with respect to  the {\em true} model $p$ decays with the number of samples $n$. $K_p$ is defined as
\begin{equation}
K_p~:=~\lim_{n\to \infty} - \frac{1}{n}\log \bP(\calA_n), \label{eqn:KP}
\end{equation}
assuming the limit exists and where $\bP$ is the product probability measure with respect to the true model $p$. We prove that the limit
in~\eqref{eqn:KP} exists in Section~\ref{sec:ee} (Corollary~\ref{thm:Jge0_gauss}). The value of $K_p$ for
different tree models $p$ provides an
indication of the relative ease  of estimating
such models. Note that both
the {\em parameters} and {\em structure}
of the model influence the magnitude of $K_p$. 

\section{Deriving the Error Exponent} \label{sec:ee}
\subsection{Crossover Rates for Mutual Information Quantities}\label{sec:crossover}
To compute $K_p$, consider first two pairs
of nodes $e,e'\in \binom{\calV}{2}$ such that  $I(p_e)> I(p_{e'})$.
We now  derive a large-deviation principle (LDP) for the {\em
crossover event of empirical mutual information quantities}
\begin{equation}
\calC_{e,e'}~:=~\{\bx^n:I(\hp_{e})\le I(\hp_{e'})\}.\label{eqn:crossover}
\end{equation}
This is an important event for the computation of $K_p$ because if
two pairs of nodes (or node pairs) $e$ and $e'$ happen to {\em crossover}, this {\em
may} lead to the event $\calA_n$ occurring (see the next subsection). We define $J_{e,e'}=J_{e,e'}(p_{e,e'})$, the  {\em
crossover rate of empirical mutual information quantities}, as
\begin{eqnarray}
J_{e,e'}~:=~\lim_{n\to \infty} - \frac{1}{n} \log \bP(\calC_{e,e'}). \label{eqn:Cee}
\end{eqnarray}
Here we remark that the following analysis does not depend on whether $e$ and $e'$ share a node. If $e$ and $e'$ do share a node, we say they are an \emph{adjacent} pair of nodes. Otherwise, we say $e$ and $e'$ are \emph{disjoint}. We also reserve the symbol $m$ to denote the total number of distinct nodes in $e$ and $e'$. Hence, $m=3$ if $e$ and $e'$ are adjacent and $m=4$ if $e$ and $e'$ are disjoint.

\begin{theorem}[LDP for Crossover of Empirical MI]\label{thm:ldp_gauss_mi}
For two  node pairs  $e, e'\in\binom{\calV}{2}$ with pdf $p_{e,e'}\in \calP_{\calN}(\bR^m)$ (for $m=3$ or $m=4$), the crossover rate for  empirical mutual information quantities is
\begin{eqnarray}
J_{e,e'}~=~ \inf_{q\in \calP_{\calN}(\bR^m) }\Big\{D(q\, ||\, p_{e,e'})  : I(q_e)=I(q_{e'})  \Big\}.
 \label{eqn:ldp_gauss}
\end{eqnarray}
The crossover rate $J_{e,e'}>0$ iff the correlation coefficients of $p_{e,e'}$ satisfy $|\rho_e|\ne |\rho_{e'}|$.
\end{theorem}
\begin{IEEEproof} ({\it Sketch})
This is an application of Sanov's Theorem~\cite[Ch.\ 3]{Deu00},    and the contraction principle~\cite[Ch.\ 3]{Den00} in large deviations theory, together with the maximum entropy principle~\cite[Ch.\ 12]{Cov06}. We remark that the proof is different from the corresponding result in \cite{Tan09}. See Appendix~\ref{app:ldp_mi}.
\end{IEEEproof}
Theorem~\ref{thm:ldp_gauss_mi} says that in order to compute the
crossover rate $J_{e,e'}$, we can restrict our attention
to a problem that involves  only an optimization over Gaussians, which is a finite-dimensional optimization problem.

\begin{figure}
\centering
\begin{picture}(200,60)
\put(0,50){\line(1,-1){50}}
\put(150,00){\line(1,1){50}}
\put(50,00){\line(1,0){100}}
\linethickness{0.5mm}
\put(50,00){\line(1,0){50}}
\put(0,50){\circle*{8}}
\put(50,0){\circle*{8}}
\put(100,0){\circle*{8}}
\put(150,0){\circle*{8}}
\put(200,50){\circle*{8}}
\put(100,60){\makebox (0,0){$e'\notin\calE_P$}}
\put(90,30){\makebox (0,0){$e\in\Path(e';\calE_P)$}}
\linethickness{0.005mm}
\multiput(0,50)(4,0){50}{\line(1,0){2}}
\put(65,23){\vector(0,-1){20}}
\end{picture} 
\caption{If the error event  occurs during the learning process, an edge $e\in\Path(e';\calE_p)$ is replaced by a non-edge $e'\notin\calE_p$ in the original model. We identify the crossover event that has the minimum rate $J_{e,e'}$ and its rate is $K_p$.}
\label{fig:replace}
\end{figure}
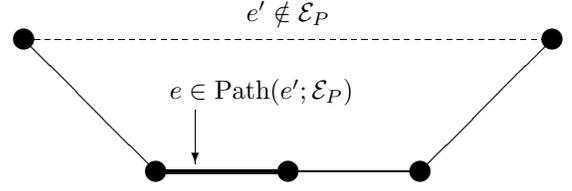

\subsection{Error Exponent for Structure Learning} \label{subsec:ee}
We now relate the set of crossover rates $\{J_{e,e'} \}$ over all the node pairs $e,e'$  to the error exponent $K_p$, defined in~\eqref{eqn:KP}.  The primary idea behind this computation is the following: We consider a fixed non-edge $e'\notin\calE_p$ in the true tree $T_p$ which may be erroneously selected during learning process. Because of the global tree constraint, this non-edge $e'$ must \emph{replace} some edge along its unique path in the original model. We only need to consider a single such crossover event because $K_p$ will be larger if there are multiple crossovers (see formal proof in~\cite{Tan09}). Finally, we identify the crossover event that has the minimum rate. See Fig.~\ref{fig:replace} for an illustration of this intuition.

\begin{theorem}[Exponent as a Crossover Event~\cite{Tan09}]  \label{thm:dom}
The error exponent for  structure learning of tree-structured Gaussian graphical models, defined in~\eqref{eqn:KP}, is given as
\begin{equation}
K_p  ~=~ \min_{e'\notin \calE_p}  \min_{e\in \Path(e';\calE_p)} \,\,
J_{e,e'} ,\label{eqn:Jfinal2}
\end{equation}
where $\Path(e';\calE_p)\subset \calE_p$ is the unique path joining the  nodes in $e'$ in the original tree $T_p= (\calV, \calE_p)$.
\end{theorem}
This theorem implies that  the {\em dominant error tree}~\cite{Tan09},
which is the asymptotically  most-likely estimated error tree under the
error event $\calA_n$,  differs from the true tree $T_p$ in
exactly   one  edge. Note that in order to compute the error exponent $K_p$ in~\eqref{eqn:Jfinal2},
we need to compute at most $\diam(T_p)(d-1)(d-2)/2$ crossover rates,
where $\diam(T_p)$ is the diameter of   $T_p$. Thus,
this is a significant reduction in the complexity of computing $K_p$
as compared to performing an exhaustive search over all possible error
events which requires a total of $\calO(d^{d-2})$
computations~\cite{West:book} (equal to the number of spanning trees
with $d$ nodes).%

In addition, from the result in Theorem~\ref{thm:dom}, we can derive
conditions to ensure that $K_p>0$ and hence for the error
probability to decay exponentially.
\begin{corollary}[Condition for Positive Error Exponent]\label{thm:Jge0_gauss}
The error probability $\bP(\calA_n)$ decays exponentially, \ie, $K_p>0$ iff  $\bSigma $   has full rank and  $T_p$ is not a forest  (as was assumed in Section~\ref{sec:prelims}).
\end{corollary}
\begin{IEEEproof} 
See Appendix~\ref{app:greaterzero} for the proof.
\end{IEEEproof}

The above result provides necessary and sufficient conditions for
the error exponent $K_p$ to be positive, which implies exponential decay of the
error probability in $n$, the number of samples. Our goal now is to analyze the influence of
structure and parameters of the Gaussian distribution $p$ on the \emph{magnitude} of the
error exponent $K_p$. Such an exercise requires a closed-form
expression for $K_p$, which in turn, requires a closed-form
expression for the crossover rate $J_{e,e'}$. However, the
crossover rate, despite having an exact expression
in~\eqref{eqn:ldp_gauss}, can only be found numerically, since the
optimization is  non-convex (due to the highly nonlinear equality constraint $
I(q_e)=I(q_{e'})$). Hence, we provide an approximation to the
crossover rate in the next section  which is tight in the so-called
very noisy learning regime.

\section{Euclidean Approximations} \label{sec:euc_gauss}
In this section, we use an approximation that only considers parameters of Gaussian tree models that are ``hard'' for learning. There are three reasons for doing this. Firstly, we expect parameters which result in easy problems to have large error exponents and so the structures can be learned  accurately from a moderate number of samples. Hard problems thus lend much more insight into when and how errors occur. Secondly, it allows us to approximate the intractable problem in~\eqref{eqn:ldp_gauss} with an intuitive, closed-form expression. Finally, such an approximation allows us to compare the relative ease of learning various tree structures in the subsequent sections.

Our analysis is based on Euclidean information theory~\cite{Bor08}, which we exploit to approximate the crossover rate $J_{e,e'}$ and the error exponent $K_p$,  defined in~\eqref{eqn:Cee} and~\eqref{eqn:KP}  respectively. The key idea is to impose suitable ``noisy'' conditions on  $p_{e,e'}$ (the joint pdf on node pairs  $e$ and $e'$) so as to enable us to relax the non-convex optimization problem in~\eqref{eqn:ldp_gauss} to a convex program.

\begin{definition}[$\epsilon$-Very Noisy Condition]
The joint pdf $p_{e,e'}$ on node pairs $e$ and $e'$ is said
to satisfy the {\em $\epsilon$-very noisy condition} if the
correlation coefficients on  $e$ and $e'$ satisfy
$
||\rho_e|-|\rho_{e'}||<\epsilon.  
$
\end{definition}
By continuity of the mutual information in the correlation coefficient, given any fixed $\epsilon$  and  $ \rho_e$, there
exists a $ \delta=\delta(\epsilon,\rho_e)>0$ such that
$|I(p_e)-I(p_{e'})|<\delta$, which means that if $\epsilon$ is
small, it is difficult to distinguish which node pair $e$ or $e'$
has the larger mutual information given the samples $\bx^n$. Therefore the ordering of the empirical mutual information quantities $I(\hp_e)$ and $I(\hp_{e'})$ may be incorrect. Thus, if $\epsilon$ is small, we are in the very noisy learning regime, where learning is difficult.

To perform our analysis, we  recall from Verdu~\cite[Sec.\ IV-E]{Verdu02} that we can
bound the KL-divergence between two zero-mean  Gaussians with covariance matrices
 $\bSigma_{e,e'}+\bDelta_{e,e'}$ and $\bSigma_{e,e'}$   as
\begin{equation}
 D(\calN(\bzero, \bSigma_{e,e'}+\bDelta_{e,e'}) || \calN(\bzero, \bSigma_{e,e'})) \!\le\!
\frac{ \|\bSigma_{e,e'}^{-1} \bDelta_{e,e'} \|_F^2}{4}, \label{eqn:gauss_app_obj}
 \end{equation}
where $\|\mathbf{M}\|_F$ is the Frobenius norm of the matrix $\mathbf{M}$. Furthermore, the inequality in~\eqref{eqn:gauss_app_obj} is tight when the perturbation matrix $\bDelta_{e,e'}$ is small. More precisely, as the ratio of the singular values
$
\frac{\sigma_{\max}(\bDelta_{e,e'} ) }{  \sigma_{\min} (\bSigma_{e,e'})} 
$
tends to zero, the inequality in \eqref{eqn:gauss_app_obj} becomes tight. To convexify the problem, we also perform a linearization of the nonlinear constraint set in \eqref{eqn:ldp_gauss} around the unperturbed covariance matrix $\bSigma_{e,e'}$. This involves taking the derivative of the mutual information with respect to the covariance matrix in the Taylor expansion. We denote this derivative as $\nabla_{\bSigma_e}I(\bSigma_e)$ where  $I(\bSigma_e)=I(\calN(\0, \bSigma_e))$ is the mutual information between the two random variables of the Gaussian joint pdf $p_{e}=\calN(\0,\bSigma_e)$. We now define the {\em linearized constraint set}
of~\eqref{eqn:ldp_gauss} as the affine subspace
\begin{align}
L_{\bDelta}(p_{e,e'})  &:=  \{\bDelta_{e,e'} \in \bR^{m\times m}   :     I(  \bSigma_e)+  \left\langle \nabla_{\bSigma_e} I(\bSigma_e),\bDelta_e \right\rangle \nn\\
&= I(  \bSigma_{e'}) + \langle \nabla_{\bSigma_{e'}} I(\bSigma_{e'}),\bDelta_{e'} \rangle\},\label{eqn:constraint_set}
\end{align}
where $\bDelta_e\in\bR^{2\times 2}$ is the sub-matrix of $\bDelta_{e,e'}\in\bR^{m\times m}$ ($m=3$ or 4) that corresponds to the covariance matrix of the node pair $e$. We also define the {\em approximate crossover rate} of $e$ and $e'$ as
the minimization of the quadratic in \eqref{eqn:gauss_app_obj} over
the affine subspace $L_{\bDelta}(p_{e,e'})$ defined in
\eqref{eqn:constraint_set}:
\begin{equation}
\tilJ_{e,e'} ~:=~ \min_{ \bDelta_{e,e'}\in L_{\bDelta}(p_{e,e'})}\,\, \, \frac{1}{4}
{\|\bSigma_{e,e'}^{-1} \bDelta_{e,e'} \|_F^2} .
\label{eqn:tilJee}
\end{equation}
Eqn.~\eqref{eqn:tilJee} is a {\em convexified}
 version of the original optimization  in~\eqref{eqn:ldp_gauss}.
This problem is not only much easier to solve, but  also provides  key
 insights as to when and how errors occur when learning the structure. We now define an additional  information-theoretic quantity  before stating the Euclidean approximation.
\begin{definition}[Information Density]
Given a pairwise joint pdf $p_{i,j}$ with marginals $p_i$ and $ p_j$, the {\em information density} denoted by $s_{i,j}:\bR^2 \rightarrow \bR$, is defined as 
\begin{equation}
s_{i,j}(x_i,x_j) :=\log \frac{p_{i,j}(x_i,x_j)}{p_i (x_i)p_j(x_j)}.
\end{equation}
\end{definition}
Hence, for each pair of variables $x_i$ and $x_j$, its associated information density $s_{i,j}$ is a random variable whose
expectation is  the mutual information of $x_i$ and $x_j$, \ie, $\bE[s_{i,j}]=I(p_{i,j})$. 


\begin{theorem}[Euclidean Approx.\ of Crossover Rate]\label{thm:euc_gauss}
The approximate crossover rate for the empirical mutual information
quantities, defined in \eqref{eqn:tilJee}, is given by
\begin{eqnarray}
\tilJ_{e,e'} = \frac{(\bE[s_{e'}-s_{e}])^2}{2\,
\var(s_{e'}-s_{e})} = \frac{(I(p_{e'})-I(p_{e}))^2}{2\,
\var(s_{e'}-s_{e}) }.\label{eqn:Jee_gauss} 
\end{eqnarray}
In addition, the approximate error exponent corresponding to $\tilJ_{e,e'}$ in \eqref{eqn:tilJee} is given by
\begin{equation}
 \tilK_p~=~\min_{e'\in\calE_p} \min_{e\in\Path(e';\calE_p)} \tilJ_{e,e'}. \label{eqn:tilKp}
\end{equation}
\end{theorem}
\begin{IEEEproof}
The proof  involves solving the  least squares problem in~\eqref{eqn:tilJee}. See Appendix~\ref{app:Jee}.
\end{IEEEproof}
We have obtained a closed-form expression for the approximate
crossover rate $\tilJ_{e,e'}$ in~\eqref{eqn:Jee_gauss}. It is
proportional to the square of the difference between the mutual
information quantities. This corresponds
  to our intuition -- that if $I(p_e)$ and $I(p_{e'})$ are relatively
well separated ($I(p_e)\gg I(p_{e'})$) then the rate $\tilJ_{e,e'}$  is
large. In addition, the SNR is also weighted by the inverse variance
of the difference of the information densities $s_e- s_{e'}$. If the variance is large, then we are uncertain
about the estimate $I(\hp_e)-I(\hp_{e'})$, thereby reducing the
rate. Theorem~\ref{thm:euc_gauss} illustrates how {\em parameters} of
Gaussian tree models affect the crossover rate.
 In the sequel, we limit our analysis to the very noisy
 regime where the above expressions apply.



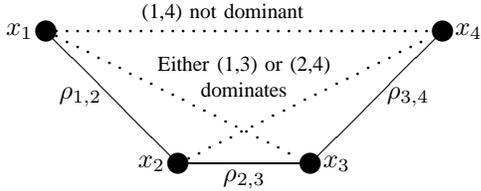
\begin{figure}
\centering
\begin{picture}(160,60)
\put(0,50){\circle*{8}}
\put(50,0){\circle*{8}}
\put(100,0){\circle*{8}}
\put(150,50){\circle*{8}}
\put(0,50){\line(1,-1){50}}
\put(50,0){\line(1,0){50}}
\put(100,0){\line(1,1){50}}
\multiput(00,50)(4,0){37}{\circle*{1}}
\multiput(0,50)(4,-2){24}{\circle*{1}}
\multiput(50,0)(4,2){24}{\circle*{1}}
\put(-10,50){\makebox (0,0){$x_1$}}
\put(40,0){\makebox (0,0){$x_2$}}
\put(110,0){\makebox (0,0){$x_3$}}
\put(160,50){\makebox (0,0){$x_4$}}
\put(13,25){\makebox (0,0){$\rho_{1,2}$}}
\put(75,-6){\makebox (0,0){$\rho_{2,3}$}}
\put(137,25){\makebox (0,0){$\rho_{3,4}$}}
\put(67,57){\makebox (0,0){{\footnotesize (1,4) not dominant}}}
\put(75,37){\makebox (0,0){{\footnotesize Either (1,3) or (2,4)}}}
\put(75,28){\makebox (0,0){{\footnotesize dominates}}}
\end{picture}
\caption{Illustration of correlation decay in a Markov chain. By Lemma~\ref{thm:mono}(b), only the node
pairs  $(1,3)$ and $(2,4)$ need to be considered for computing the error exponent $\tilK_p$. By  correlation decay, the node pair
$(1,4)$ will not be mistaken as a true edge by the  estimator because its distance, which is equal to 3, is longer than either  $(1,3)$ or $(2,4)$, whose distances are equal to 2.  } \label{fig:chain}
\end{figure}

\section{Simplification of the Error Exponent} \label{sec:simplifyKp}
In this section, we exploit the properties of the approximate crossover rate in~\eqref{eqn:Jee_gauss} to significantly reduce the complexity in finding the error exponent $\tilK_p$ to $\calO(d)$. As a motivating example, consider the Markov chain in Fig.~\ref{fig:chain}. From our analysis to this point, it appears that, when computing the approximate error exponent $\tilK_p$ in~\eqref{eqn:tilKp}, we have to consider all possible replacements between the non-edges $(1,4)$, $(1,3)$ and $(2,4)$ and the true edges along the unique  paths connecting these non-edges. For example, $(1,3)$ might be mistaken as a true edge, replacing either $(1,2)$ or $(2,3)$.

We will prove that, in fact, to compute $\tilK_p$ we can ignore the possibility that longest non-edge $(1,4)$ is mistaken as a true edge, thus reducing the number of computations for the approximate crossover rate $\tilJ_{e,e'}$. The key to this result is the exploitation of \emph{correlation decay}, \ie,  the decrease in the absolute value of the correlation coefficient between two nodes as the \emph{distance} (the number of edges along the path between two nodes) between them increases. This follows from the Markov property: 
\begin{equation}
\rho_{e'} = \prod_{e \in \Path(e';\calE_p)} \rho_e, \quad \forall
e'\notin \calE_p. \label{eqn:product_corr}
\end{equation}
For example, in Fig.~\ref{fig:chain}, $|\rho_{1,4}|\le\min\{|\rho_{1,3}|,|\rho_{2,4}|\}$ and because of this, the following lemma implies that $(1,4)$ is less likely to be mistaken as a true edge than $(1,3)$ or $(2,4)$. 


\begin{figure}
  \centering
  \subfigure 
  {
      \includegraphics[width=2.5in]{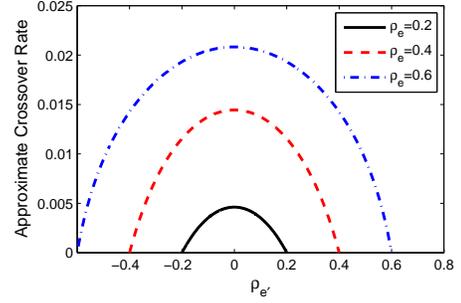}
      \label{fig:hist1}
  }      \hspace{.3in}
  \subfigure 
  {
      \includegraphics[width=2.5in]{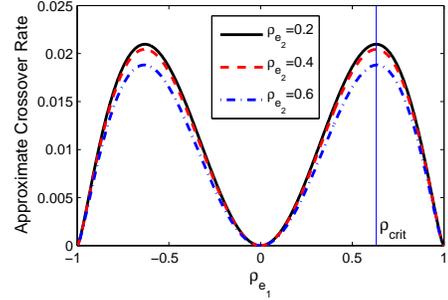}
      \label{fig:hist2}
  }
\hspace{.3in}  
 {
      \includegraphics[width=2.5in]{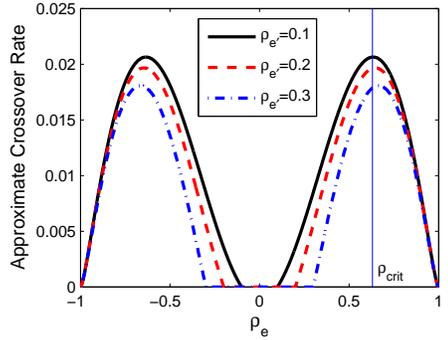}
      \label{fig:hist3}
  }  
   \caption{Illustration of the properties of $\tilJ(\rho_e,\rho_{e'})$ in Lemma~\ref{thm:mono}.  $\tilJ(\rho_e, \rho_{e'})$ is  decreasing in $|\rho_{e'}|$  for fixed $\rho_e$ (top) and $\tilJ(\rho_{e_1}, \rho_{e_1} \rho_{e_2})$ is increasing in $|\rho_{e_1}|$  for fixed $\rho_{e_2}$ if $|\rho_{e_1}|<\rho_{\mathrm{crit}}$ (middle). Similarly, $\tilJ(\rho_e,\rho_{e'})$ is increasing in $|\rho_e|$ for fixed $\rho_{e'}$ if  $|\rho_{e}|<\rho_{\mathrm{crit}}$ (bottom). }
  \label{fig:prop}
\end{figure}

It is easy to verify that the crossover rate $\tilJ_{e,e'}$ in~\eqref{eqn:Jee_gauss} depends {\em only} on the correlation coefficients $\rho_e$ and $\rho_{e'}$ and not the variances $\sigma_i^2$. Thus, without loss of generality, we assume that all random variables have unit variance (which is still unknown to the learner) and to make the dependence clear, we now write  $\tilJ_{e,e'}=\tilJ(\rho_e, \rho_{e'})$. Finally define $\rho_{\mathrm{crit}}:=0.63055$. 
\begin{lemma}[Monotonicity of $\tilJ(\rho_e, \rho_{e'})$] \label{thm:mono}
$\tilJ(\rho_e, \rho_{e'})$, derived in \eqref{eqn:Jee_gauss}, has the following properties:
\begin{enumerate}
\item[(a)] $\tilJ(\rho_e, \rho_{e'})$ is an even function of both $\rho_e$ and $\rho_{e'}$.
\item[(b)] $\tilJ(\rho_e, \rho_{e'})$ is monotonically {\it decreasing} in $|\rho_{e'}|$ for fixed $\rho_e \in (-1,1)$.
\item[(c)] Assuming that $|\rho_{e_1}|<\rho_{\mathrm{crit}}$, then $\tilJ(\rho_{e_1}, \rho_{e_1} \rho_{e_2})$ is monotonically {\it increasing} in $|\rho_{e_1}|$ for fixed $\rho_{e_2}$.
\item[(d)] Assuming that $|\rho_{e}|<\rho_{\mathrm{crit}}$, then $\tilJ(\rho_{e}, \rho_{e'})$ is monotonically {\it increasing} in $|\rho_{e}|$ for fixed $\rho_{e'}$.
\end{enumerate}
See Fig.~\ref{fig:prop} for an illustration of the properties of $\tilJ(\rho_e, \rho_{e'})$.  
\end{lemma}

\begin{IEEEproof} ({\it Sketch})
Statement (a) follows from~\eqref{eqn:Jee_gauss}. We prove  (b) by showing that $\partial\tilJ(\rho_e,
\rho_{e'})/\partial |\rho_{e'}|\le 0 $ for all $|\rho_{e'}| \le
|\rho_{e}|$. Statements (c) and (d) follow similarly. See Appendix~\ref{app:mono}
for the details.
\end{IEEEproof}

Our intuition about correlation decay is substantiated by Lemma~\ref{thm:mono}(b), which implies that for the example in Fig.~\ref{fig:chain}, $\tilJ(\rho_{2,3}, \rho_{1,3}) \le \tilJ(\rho_{2,3},\rho_{1,4}),$ since  $|\rho_{1,4}|\le  |\rho_{1,3}|$ due to Markov property on the chain~\eqref{eqn:product_corr}. Therefore, $\tilJ(\rho_{2,3},\rho_{1,4})$ can be ignored in the minimization to find $\tilK_p$ in~\eqref{eqn:tilKp}. Interestingly while Lemma~\ref{thm:mono}(b) is a statement about correlation decay, Lemma~\ref{thm:mono}(c) states that the absolute strengths of the correlation coefficients also influence the magnitude of the crossover rate. 

From Lemma~\ref{thm:mono}(b) (and the above motivating example in  Fig.~\ref{fig:chain}), finding the approximate error exponent $\tilK_p$ now reduces to finding the minimum crossover rate only over {\em triangles} ($(1,2,3)$ and $(2,3,4)$) in the tree as shown in Fig.~\ref{fig:chain}, \ie, we only need to consider $\tilJ(\rho_e, \rho_{e'})$ for \emph{adjacent} edges.  


\begin{corollary}[Computation of $\tilK_p$] Under the very noisy
learning regime, the error exponent $\tilK_p$ is 
\begin{equation}
\widetilde{K}_p ~=~ \min_{e_i,e_j \in \calE_p, e_i\sim e_j }
W(\rho_{e_i},\rho_{e_j}), \label{eqn:triangle}\end{equation} where
$e_i\sim e_j$ means that the edges $e_i$ and $e_j$ are adjacent
and  the   weights are defined as \begin{eqnarray}
W(\rho_{e_1},\rho_{e_2})\! := \!\min  \left\{\!    \tilJ(\rho_{e_1},
\rho_{e_1}\rho_{e_2}) ,
  \tilJ(\rho_{e_2}, \rho_{e_1}\rho_{e_2})\!  \right\}.  \label{eqn:wts}
\end{eqnarray}
\end{corollary}


If we  carry out the computations in~\eqref{eqn:triangle} independently,
the complexity is $\calO(d \Deg_{\max})$, where $\Deg_{\max}$ is the
maximum degree of the nodes in the tree graph. Hence, in the worst case, the complexity
is $\calO(d^2)$, instead of $\calO(d^3)$ if~\eqref{eqn:tilKp} is
used. We can, in fact, reduce the number of computations to $\calO(d)$.

\begin{proposition}[Complexity in computing $\tilK_p$]\label{lem:facttilK}
The approximate error exponent $\tilK_p$, derived in \eqref{eqn:tilKp}, can be computed in linear time ($d-1$ operations) as
\begin{equation}
\tilK_p ~=~ \min_{e\in \calE_p} \,\, \tilJ(\rho_e,\rho_e\rho_{e}^*), \label{eqn:tilKp_simple}
\end{equation}
where the maximum correlation coefficient on  the edges adjacent to $e\in \calE_p$ is defined as
\begin{equation}
\rho_e^* ~:=~ \max\{|\rho_{\tilde{e}}|: \tilde{e}\in\calE_p, \tilde{e}\sim e\}.\label{eqn:rhoestar}
\end{equation}
\end{proposition}
\begin{IEEEproof}
By Lemma~\ref{thm:mono}(b) and the definition of $\rho_e^*$, we obtain the  smallest crossover rate associated to edge $e$. We obtain the approximate error exponent $\tilK_p$ by
minimizing over all edges $e\in \calE_p$ in~\eqref{eqn:tilKp_simple}.
\end{IEEEproof}
Recall that $\diam(T_p)$ is the diameter of $T_p$. The computation of $K_p$ is reduced significantly from
$\calO(\diam(T_p)d^2)$ in~\eqref{eqn:Jfinal2} to $\calO(d)$. Thus,
there is a further reduction in the complexity to estimate the error
exponent $K_p$ as compared to exhaustive search which requires
$\calO(d^{d-2})$ computations. This simplification only holds
  for Gaussians under the very noisy regime.

\def\calpath{{\mathcal{S}}}
\def\calpathd{{\mathcal{S}^d}}
\def\calpathdminus{{\mathcal{S}^{d-1}}}

\section{Extremal Structures for Learning} \label{sec:structure}
In this  section, we study the influence of graph structure on the approximate error exponent $\tilK_p$  using the concept of correlation decay and the properties of the crossover rate $\tilJ_{e,e'}$ in Lemma~\ref{thm:mono}. We have already discussed the connection between the error exponent and  correlation decay. We also proved that non-neighbor node pairs which have shorter distances are more likely to be mistaken as edges by the ML estimator. Hence, we expect that a tree $T_p$ which contains non-edges with shorter distances to be ``harder" to learn (i.e., has a smaller error exponent $\tilK_p$) as compared to a tree which contains non-edges with longer distances. In subsequent subsections, we formalize this intuition in
terms of the diameter of the tree $\diam(T_p)$, and show that the extremal trees, in terms of their diameter, are also extremal trees for learning. We also analyze the effect of changing the size of the tree on the error exponent. 

From the Markov property in~\eqref{eqn:product_corr}, we see that  for a  Gaussian tree distribution, the set of correlation coefficients fixed on the edges of the tree, along with the structure $T_p$, are  sufficient statistics and they completely characterize $p$. Note that this parameterization neatly decouples the structure from the correlations. We use this fact to study the influence of changing the structure $T_p$ while keeping the set of correlations    on the edges fixed.\footnote{Although the set of correlation coefficients on the
edges is fixed, the elements in this set can be arranged in different ways on the edges of the tree. We formalize this concept in~\eqref{eqn:set_perm}.} Before doing so, we provide a  review of some basic graph theory.

\subsection{Basic Notions in Graph Theory} \label{sec:basics_graphs}
\begin{definition}[Extremal Trees in terms of Diameter] Assume that $d>3$. Define the {\em extremal trees} with $d$ nodes in terms of the tree diameter $\diam:\calT^d\rightarrow \{2,\ldots, d-1\}$ as
\begin{equation}
T_{\max}(d)\!:=\!\argmax_{T\in\calT^d}\diam(T), \,\,\,
T_{\min}(d)\!:=\!\argmin_{T\in\calT^d}\diam(T), \label{eqn:extremal}
\end{equation}
Then it is clear that the two extremal structures, the \emph{chain} (where there is a simple path passing through all nodes and edges exactly once) and the \emph{star} (where there is one central node)  have the largest and smallest diameters respectively, \ie,  $ T_{\max}(d) = T_{\mathrm{chain}}(d), $ and  $T_{\min}(d) =
T_{\mathrm{star}}(d).$
\end{definition}
\begin{figure}
\centering
\begin{tabular}{cc}
\begin{picture}(100,55)
\put(0,0){\line(1,0){100}}
\put(50,0){\line(0,1){50}}
\put(0,50){\line(1,0){100}}
\put(0,0){\circle*{8}}
\put(50,0){\circle*{8}}
\put(0,50){\circle*{8}}
\put(100,0){\circle*{8}}
\put(50,50){\circle*{8}}
\put(100,50){\circle*{8}}
\put(25,7){\makebox (0,0){$e_1$}}
\put(75,7){\makebox (0,0){$e_2$}}
\put(45,25){\makebox (0,0){$e_3$}}
\put(25,42){\makebox (0,0){$e_4$}}
\put(75,42){\makebox (0,0){$e_5$}}
\end{picture}  & \hspace{.25in}
\begin{picture}(100,55)
\put(0,0){\line(1,0){100}}
\put(0,0){\line(0,1){50}}
\put(50,0){\line(0,1){50}}
\put(0,50){\line(1,-1){50}}
\put(50,50){\line(1,-1){50}}
\put(0,0){\circle*{8}}
\put(50,0){\circle*{8}}
\put(0,50){\circle*{8}}
\put(100,0){\circle*{8}}
\put(50,50){\circle*{8}}
\put(7,10){\makebox (0,0){$1$}}
\put(11,50){\makebox (0,0){$2$}}
\put(58,10){\makebox (0,0){$3$}}
\put(61,50){\makebox (0,0){$4$}}
\put(100,10){\makebox (0,0){$5$}}
\end{picture}\vspace{.05in}\\
(a)&(b)
\end{tabular}
\caption{(a): A graph $G$. (b): The line graph $H=\calL(G)$ that corresponds to $G$ is the graph whose vertices are the edges of $G$ (denoted as $e_i$) and there is an edge between any two vertices $i$ and $j$ in $H$ if the corresponding edges in $G$ share a node.}
\label{fig:linegraph}
\end{figure}
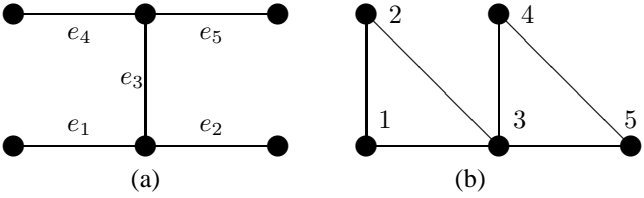

\begin{definition}[Line Graph] \label{def:line}
The {\em line graph}~\cite{West:book} $H$ of a graph $G$, denoted by $H=\calL(G)$, is one in which, roughly speaking, the vertices and edges of $G$ are interchanged. More precisely, $H$ is the  undirected graph whose vertices are the edges of $G$ and there is an edge between any two vertices in the line graph if the corresponding edges in $G$
have a common node, \ie, are adjacent. See Fig.~\ref{fig:linegraph} for a graph $G$ and its associated line graph $H$.
\end{definition}

\subsection{Formulation: Extremal  Structures for Learning}
We now formulate the problem of finding the best and worst tree
structures for learning and also the distributions
associated with them. At a high level, our strategy  involves two distinct steps. Firstly and primarily, we find the \emph{structure} of the optimal distributions in Section~\ref{sec:results_bw}. It turns out that the optimal structures that maximize and minimize the exponent are the Markov chain (under some conditions on the correlations) and the star respectively and these are the extremal structures in terms of the diameter. Secondly, we optimize over the \emph{positions} (or placement) of the correlation coefficients on the edges  of the optimal structures.

Let $\rhobf:=[ \rho_1, \rho_2, \ldots , \rho_{d-1}]$ be a \emph{fixed} vector of feasible\footnote{We do not allow any of the correlation coefficient to be zero because otherwise, this would result in $T_p$ being a  forest.}  correlation coefficients, \ie, $\rho_i\in (-1, 1)\setminus \{0\}$ for all $i$. For a tree, it follows from~\eqref{eqn:product_corr} that if $\rho_i$'s are the correlation coefficients on the edges, then $|\rho_i|< 1$ is a necessary and sufficient condition to ensure that $\bSigma\succ 0$.  Define $\bm{\Pi}_{d-1}$ to be the group of permutations of order $d-1$, hence elements in $\bm{\Pi}_{d-1}$  are permutations of a given ordered set with cardinality $d-1$. Also denote the set of tree-structured, $d$-variate Gaussians  which have unit variances at all nodes and $\rhobf$ as the correlation coefficients on the edges in some order as $\calP_{\calN}(\bR^d,\calT^d; \rhobf)$. Formally,
\begin{align}
&\calP_{\calN}(\bR^d,\calT^d; \rhobf):= \big\{ p(\bx) \!=\! \calN(\bx; \0, \bSigma) \!\in\! \calP_{\calN}(\bR^d, \calT^d):  \nn\\ &\bSigma(i,i)=1,\forall \,  i\in \calV,\exists\, \bm{\pi}_p\in
\bm{\Pi}_{d-1}: \bsigma_{\calE_p} = \bm{\pi}_p(\rhobf)
\big\},\label{eqn:set_perm}
\end{align}
where $\bsigma_{\calE_p} :=[ \bSigma(i,j) : (i,j)\in\calE_p]$ is the length-$(d-1)$ vector consisting of
the covariance elements\footnote{None of the elements in $\bSigma$ are allowed to be zero because $\rho_i\ne 0$ for every $i\in\calV$ and the Markov property in~\eqref{eqn:product_corr}.} on the edges (arranged in lexicographic order) and $\bm{\pi}_p(\rhobf)$ is the permutation  of $\rhobf$ according to $\bm{\pi}_p$. The tuple $(T_p, \bm{\pi}_p, \rhobf)$ uniquely parameterizes a Gaussian tree
distribution with unit variances. Note that we can regard the permutation $\bpi_p$ as a nuisance parameter for solving the optimization for the best structure given $\rhobf$. Indeed, it can happen that there are different $\bpi_p$'s such that the error exponent $\tilK_p$ is the same. For instance, in a star graph, all permutations $\bpi_p$ result in the same exponent. Despite this, we show that extremal tree \emph{structures} are invariant to the specific choice of $\bpi_p$ and  $\rhobf$.

For distributions in the set $\calP_{\calN}(\bR^d,\calT^d; \rhobf)$,
our goal is   to find the best (easiest to learn) and the worst   (most difficult to learn) distributions for
learning. Formally, the optimization problems for the best
and worst distributions for learning are given by
\begin{align}
p_{\max, \rhobf} &~:=~ \argmax_{p\in \calP_{\calN}(\bR^d,\calT^d; \rhobf)}\tilK_p, \label{eqn:max_prob}\\
p_{\min, \rhobf} &~:=~ \argmin_{p\in \calP_{\calN}(\bR^d,\calT^d; \rhobf)}\tilK_p. \label{eqn:min_prob}
\end{align}
Thus, $p_{\max, \rhobf}$  (resp.\ $p_{\min, \rhobf}$) corresponds to the Gaussian tree model which has the largest (resp.\ smallest) approximate error exponent. 

\subsection{Reformulation as Optimization  over Line Graphs}

Since the number of  permutations $\bpi$ and number of
spanning trees are prohibitively large, finding the
optimal distributions cannot be done through
a brute-force search unless $d$ is small. Our main idea in this section is to use the notion of line graphs to simplify the problems in~\eqref{eqn:max_prob} and~\eqref{eqn:min_prob}. In subsequent sections, we  identify the extremal tree structures before identifying the precise best and worst distributions.

Recall that the approximate error exponent $\tilK_p$ can be expressed in terms of the weights $W(\rho_{e_i} ,\rho_{e_j})$   between two adjacent edges $e_i,e_j$ as in~\eqref{eqn:triangle}. Therefore, we can write the extremal distribution in~\eqref{eqn:max_prob} as
\begin{equation}
p_{\max, \rhobf} = \argmax_{p\in\calP_{\calN}(\bR^d,\calT^d;
\rhobf) } \min_{e_i,e_j\in \calE_p, e_i\sim e_j} W( \rho_{e_i},
\rho_{e_j})  \label{eqn:pmaxrho} .\end{equation}
Note that in~\eqref{eqn:pmaxrho}, $\calE_p$ is the edge set of a weighted graph whose edge weights are given by $\rhobf$.  Since the weight is between two edges, it is more  convenient to consider line graphs  defined in
Section~\ref{sec:basics_graphs}.

We now transform the intractable optimization problem in~\eqref{eqn:pmaxrho} over the set of trees to an optimization problem over all the set of  line graphs:
\begin{equation}
p_{\max, \rhobf} = \argmax_{ p\in\calP_{\calN}(\bR^d,\calT^d; \rhobf)}  \min_{\substack{  (i,j)\in H, H = \calL(T_p)}} W(\rho_i,\rho_j), \label{eqn:pmax_line}
\end{equation}
and $W(\rho_i,\rho_j)$ can be considered as an edge weight between
nodes $i$ and $j$ in a  weighted line graph $H $. Equivalently,~\eqref{eqn:min_prob}  can also
be written as in~\eqref{eqn:pmax_line} but with the $\argmax$ replaced by an $\argmin$.

\subsection{Main Results: Best and Worst Tree Structures} \label{sec:results_bw}
In order to solve~\eqref{eqn:pmax_line}, we need to
characterize the set of line graphs of  spanning trees
$\calL(\calT^d)=\{\calL(T):T\in\calT^d\}$. This has been studied before~\cite[Theorem 8.5]{Harary72}, but the set $\calL(\calT^d)$ is
nonetheless still very complicated. Hence, solving~\eqref{eqn:pmax_line} directly is 
intractable. Instead, our strategy now is to identify  the
{\em structures} corresponding to the optimal distributions,
$p_{\max, \rhobf}$ and $p_{\min, \rhobf}$ by exploiting the
monotonicity of $\tilJ(\rho_e,\rho_{e'})$ given in Lemma~\ref{thm:mono}.

\begin{theorem}[Extremal Tree Structures] \label{thm:graph_topo}
The  tree structure that minimizes the   approximate error exponent $\tilK_p$ in~\eqref{eqn:min_prob}  is given by
\begin{equation}
T_{p_{\min, \rhobf}}  = T_{\mathrm{star}}(d), \label{eqn:worstresult}
\end{equation}
for all feasible correlation coefficient vectors $\rhobf$ with $\rho_i\in (-1,
1)\setminus\{0\}$. In addition, if $\rho_i\in (-\rho_{\mathrm{crit}}, \rho_{\mathrm{crit}})\setminus \{0\}$ (where $\rho_{\mathrm{crit}}=0.63055$), then the tree structure that maximizes the   approximate error exponent $\tilK_p$ in~\eqref{eqn:max_prob} is given by 
\begin{equation}
T_{p_{\max, \rhobf}}  = T_{\mathrm{chain}}(d),  \label{eqn:bestresult}
\end{equation}
\end{theorem}
 \begin{figure}
\centering
\begin{tabular}{cc}
\begin{picture}(100,57)
\put(0,30){\line(1,0){100}}
\put(50,0){\line(0,1){60}}
\put(0,30){\circle*{8}}
\put(50,0){\circle*{8}}
\put(50,30){\circle*{8}}
\put(50,60){\circle*{8}}
\put(100,30){\circle*{8}}
\put(60,15){\makebox (0,0){$\rho_{e_3}$}}
\put(60,45){\makebox (0,0){$\rho_{e_1}$}}
\put(75,37){\makebox (0,0){$\rho_{e_2}$}}
\put(25,37){\makebox (0,0){$\rho_{e_4}$}}
\end{picture} & \hspace{.15in}
\begin{picture}(120,70)
\put(0,30){\line(1,0){120}}
\put(0,30){\circle*{8}}
\put(30,30){\circle*{8}}
\put(60,30){\circle*{8}}
\put(90,30){\circle*{8}}
\put(120,30){\circle*{8}}
\put(15,37){\makebox (0,0){$\rho_{e_{1}}$}}
\put(45,37){\makebox (0,0){$\rho_{e_{2}}$}}
\put(75,37){\makebox (0,0){$\rho_{e_{3}}$}}
\put(105,37){\makebox (0,0){$\rho_{e_{4}}$}}
\end{picture} \\
(a) & (b)
\end{tabular}
\caption{Illustration for Theorem \ref{thm:graph_topo}: The star (a) and the  chain  (b) minimize and maximize the approximate error exponent respectively. }
\label{fig:bestworst}
\end{figure}
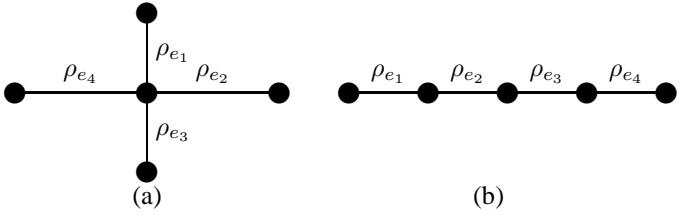
\begin{IEEEproof} ({\it Idea})
The assertion that $T_{p_{\min, \rhobf}} = T_{\mathrm{star}}(d)$  follows from the fact that all the crossover rates for the star graph are the minimum possible, hence  $\tilK_{\mathrm{star}} \le  \tilK_p$. See Appendix~\ref{app:bestworst} for the details. 
\end{IEEEproof}
See Fig.~\ref{fig:bestworst}. This theorem agrees with our intuition: for the star graph, the
nodes are strongly correlated (since its diameter is the smallest) while
in the chain, there are many weakly correlated pairs of nodes for the same set of
correlation coefficients on the edges thanks to correlation decay. Hence, it is hardest to learn
the star while it is easiest to learn the chain. It is
interesting to observe Theorem~\ref{thm:graph_topo} implies that  the extremal tree structures
$T_{p_{\max, \rhobf}}$ and $T_{p_{\min, \rhobf}}$ are {\em independent of} the correlation coefficients $\rhobf$ (if $|\rho_i|<\rho_{\mathrm{crit}}$ in the case of the star). Indeed, the experiments in Section~\ref{sec:compare_diff_trees} also suggest  that Theorem~\ref{thm:graph_topo} may likely be true for larger ranges of problems (without the constraint that $|\rho_i|<\rho_{\mathrm{crit}}$) but this remains open.

The results in~\eqref{eqn:worstresult} and~\eqref{eqn:bestresult} do not yet provide the complete solution
to $p_{\max, \rhobf}$ and $p_{\min, \rhobf}$ in~\eqref{eqn:max_prob} and~\eqref{eqn:min_prob} since there are many
possible pdfs in $\calP_{\calN}(\bR^d, \calT^d;\rhobf)$ corresponding to a fixed tree because we can rearrange the correlation coefficients along the edges of the tree in multiple ways. The only exception is  if $T_p$ is known to be a star then  there is only one  pdf in $\calP_{\calN}(\bR^d, \calT^d;\rhobf)$, and we formally state the result below.

\begin{corollary}[Most Difficult Distribution to Learn] \label{cor:worst_dist}
The Gaussian $p_{\min, \rhobf}(\bx)\!=\! \calN(\bx;\0, \bSigma_{\min,\rhobf})$ defined in~\eqref{eqn:min_prob},   corresponding to the most difficult distribution to learn for fixed $\rhobf$, has the
covariance matrix whose upper triangular elements are given as $\bSigma_{\min,\rhobf}(i,j)=\rho_i$ if $i=1, j\ne 1$ and  $\bSigma_{\min,\rhobf}(i,j)=\rho_i\rho_j$ otherwise. 
Moreover, if $|\rho_1|\ge \ldots\ge|\rho_{d-1}|$ and
$|\rho_1|<\rho_{\mathrm{crit}}=0.63055$, then  $\tilK_p$
corresponding to the star graph  can be written
explicitly as a  minimization over only two crossover rates: $
\tilK_{p_{\min,\rhobf}}=\min \{\tilJ(\rho_1, \rho_1\rho_2),
\tilJ(\rho_{d-1}, \rho_{d-1}\rho_1) \}.$
\end{corollary}
\begin{IEEEproof}
The first assertion follows from the Markov property~\eqref{eqn:product_corr} and Theorem~\ref{thm:graph_topo}. The  next result follows from Lemma~\ref{thm:mono}(c) which implies that $\tilJ(\rho_{d-1}, \rho_{d-1}\rho_1)\le  \tilJ(\rho_{k}, \rho_{k}\rho_1)$ for all $2\le k\le d-1$.
\end{IEEEproof}
In other words,  $p_{\min, \rhobf}$ is a {\it star} Gaussian
graphical model with correlation coefficients $\rho_i$ on its edges. This result can also be explained by correlation decay. In a star graph, since the distances between  non-edges are small, the estimator in~\eqref{eqn:cloptgauss} is more likely to mistake a non-edge with a true edge. It is often useful in applications to compute the minimum error exponent for a fixed vector of correlations
$\rhobf$ as it provides a lower bound of the decay rate of $\bP(\calA_n)$ for any tree
distribution with parameter vector $\rhobf$. Interestingly, we also have a  result for the easiest tree distribution to learn. 

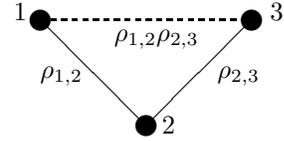
\begin{figure}
\centering
\begin{picture}(80, 40)
\put(0,40){\circle*{8}}
\put(40,0){\circle*{8}}
\put(80,40){\circle*{8}}

\put(0,40){\line(1,-1){40}}
\put(40,00){\line(1,1){40}}
 
\multiput(0,40)(4,0){20}{\line(1,0){2}}

\put(0,18){\mbox{$\rho_{1,2}$}}
\put(67,18){\mbox{$\rho_{2,3}$}}
\put(28,32){\mbox{$\rho_{1,2}\rho_{2,3}$}}

\put(-10,40){\mbox{$1$}}
\put(87,40){\mbox{$3$}}
\put(46,-3){\mbox{$2$}}

\end{picture}
\caption{If $|\rho_{1,2}|<|\rho_{2,3}|$, then the likelihood of the non-edge $(1,3)$ replacing edge $(1,2)$ would be higher than if $|\rho_{1,2}|=|\rho_{2,3}|$. Hence, the weight $W(\rho_{1,2},\rho_{2,3})$ is maximized when equality holds.}
\label{fig:rhojrhoi}
\end{figure}

\begin{corollary}[Easiest Distribution to Learn] \label{cor:best_dist}
Assume that $\rho_{\mathrm{crit}}> |\rho_1|\ge|\rho_2|\ge \ldots \ge|\rho_{d-1}|$. Then, the Gaussian $p_{\max, \rhobf}(\bx)\!=\! \calN(\bx;\0, \bSigma_{\max,\rhobf})$ defined in~\eqref{eqn:max_prob},   corresponding to the easiest distribution to learn for fixed $\rhobf$, has the covariance matrix whose upper triangular elements are $\bSigma_{\max,\rhobf}(i,i+1)=\rho_i$ for all $i=1,\ldots, d-1$ and $\bSigma_{\max,\rhobf}(i,j) =\prod_{k=i}^{j-1}\rho_{k}$ for all $j>i$.
\end{corollary}
\begin{IEEEproof}
The first assertion follows from the proof of Theorem~\ref{thm:graph_topo} in Appendix \ref{app:bestworst} and the second assertion from the Markov property in~\eqref{eqn:product_corr}. 
\end{IEEEproof}
In other words, in the regime where $|\rho_i|<\rho_{\mathrm{crit}}$, $p_{\max,\rhobf}$ is a {\it Markov chain} Gaussian graphical model with correlation coefficients arranged in increasing (or decreasing) order on its edges. We now provide some intuition for why this is so. If a particular correlation coefficient $\rho_i$ (such that $|\rho_i|<\rho_{\mathrm{crit}}$) is fixed, then the edge weight $W(\rho_i,\rho_j)$, defined in~\eqref{eqn:wts}, is maximized when $|\rho_j|=|\rho_i|$. Otherwise, if $|\rho_i|<|\rho_j|$ the event that the non-edge with correlation $\rho_i\rho_j$ replaces the edge with correlation  $\rho_i$ (and hence results in an error) has a higher likelihood than if equality holds.  Thus, correlations $\rho_i$ and $\rho_j$ that are close in terms of their absolute values should be placed closer to one another (in terms of graph distance) for the approximate error exponent to be maximized. See Fig.~\ref{fig:rhojrhoi}.

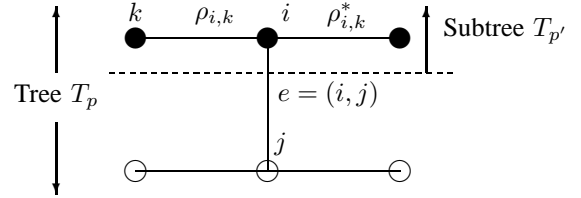
\begin{figure}
\centering
\begin{picture}(170,60)
\put(30,0){\line(1,0){100}}
\put(30,50){\line(1,0){100}}
\put(80,0){\line(0,1){50}}
\put(30,0){\circle{8}}
\put(30,50){\circle*{8}}
\put(80,0){\circle{8}}
\put(80,50){\circle*{8}}
\put(130,0){\circle{8}}
\put(130,50){\circle*{8}}
\put(87,60){\makebox (0,0){$i$}}
\put(85,10){\makebox (0,0){$j$}}
\put(30,60){\makebox (0,0){$k$}}
\put(60,58){{\makebox (0,0){$\rho_{i,k}$}}}
\put(110,58){{\makebox (0,0){$\rho_{i,k}^*$}}}
\put(103,29){\makebox (0,0){$e=(i,j)$}}
\multiput(20,37)(4,0){33}{\line(1,0){2}}
\put(140,37){\vector(0,1){25}}
\put(170,53){{\makebox (0,0){Subtree $T_{p'}$}}}
\put(0,28){{\makebox (0,0){Tree $T_p$}}}
\put(0,37){\vector(0,1){25}}
\put(0,18){\vector(0,-1){27}}
\end{picture}  
\caption{Illustration of Proposition~\ref{prop:subtree}. $T_p=(\calV, \calE_{p})$ is the original tree and $e\in\calE_p$. $T_{p'}=(\calV', \calE_{p'})$ is a subtree. The observations for learning the structure $p'$ correspond to the  shaded  nodes, the unshaded nodes correspond to unobserved variables. }
\label{fig:subgraph}
\end{figure}
\subsection{Influence of Data Dimension on Error Exponent}
 We now analyze the influence of {\em
changing} the \emph{size} of the tree on the error exponent, \ie, adding and
deleting nodes and edges while satisfying the tree constraint and
observing samples from the modified graphical model. This is of  importance in many applications. For example, in {\it sequential} problems, the learner receives data at different times and would like to update the estimate of the tree structure learned. In {\it dimensionality reduction}, the learner is required to estimate the structure of a smaller model given high-dimensional data.   Intuitively, learning
only a tree with a smaller number of nodes is easier than learning
the entire tree since there are fewer ways for errors to occur during the learning process. We prove this in the affirmative in Proposition~\ref{prop:subtree}.

Formally, we start with a $d$-variate Gaussian  $p\in\calP_{\calN}(\bR^d,\calT^d;\rhobf)$ and
consider a $d'$-variate pdf $p'\in\calP_{\calN}(\bR^{d'},\calT^{d'};\rhobf')$, obtained by marginalizing $p$ over
a subset of variables and $T_{p'}$ is the tree\footnote{Note that $T_{p'}$ still needs to satisfy the tree constraint so that the variables that are marginalized out are not arbitrary (but must be variables that form the first part of a node elimination order~\cite{Lau96}). For example, we are not allowed to marginalize out the central node of a star graph since the resulting graph would not be a tree. However, we can marginalize out any of the other nodes. In effect, we can only marginalize out nodes with degree either 1 or 2. }  associated to the distribution $p'$. Hence $d'<d$ and $\rhobf'$ is a subvector of $\rhobf$. See Fig.~\ref{fig:subgraph}. In our formulation, the only available observations are those sampled from the smaller Gaussian graphical model $p'$.

\begin{proposition}[Error Exponent of Smaller Trees] \label{prop:subtree}
The approximate error exponent for learning $p'$ is at least  that of $p$, \ie, $\tilK_{p'}\ge \tilK_{p}$.
\end{proposition}
\begin{IEEEproof}
Reducing the number of adjacent edges to a fixed edge $(i,k)\in\calE_p$ as in Fig.~\ref{fig:subgraph} (where $k\in\nbd(i)\setminus\{j\}$) ensures that the maximum correlation coefficient $\rho_{i,k}^*$, defined in~\eqref{eqn:rhoestar}, does not increase. By Lemma \ref{thm:mono}(b) and~\eqref{eqn:tilKp}, the approximate error exponent $\tilK_p$ does not decrease. 
\end{IEEEproof}
Thus, lower-dimensional models are easier to learn if the set of correlation coefficients is fixed and the tree
constraint remains satisfied. This is a consequence of the fact that there are fewer crossover error events that contribute to the error exponent $\tilK_p$. 

We now consider the ``dual'' problem of adding a new edge to an existing
tree model, which results in a larger tree. We are now
provided with $(d+1)$-dimensional observations to learn the larger
tree.  More precisely, given a
$d$-variate tree Gaussian pdf $p$, we consider a $(d+1)$-variate
 pdf $p''$ such that $T_p$ is a subtree of $T_{p''}$.
Equivalently, let $\rhobf:=[\rho_{e_1}, \rho_{e_2},\ldots, \rho_{e_{d-1}} ]$ be the vector of
correlation coefficients on the edges of the graph of $p$ and let
$\rhobf'':=[\rhobf , \rho_{\mathrm{new}}]$ be that of $p''$.

By comparing the error exponents $\tilK_p$ and $\tilK_{p''}$, we can address the following question: Given a new edge
correlation coefficient $\rho_{\mathrm{new}}$, how should one adjoin this new
edge to the existing tree such that the resulting error exponent
is maximized or minimized? Evidently, from Proposition~\ref{prop:subtree},
it is not possible to increase the error exponent by growing
the tree but can we devise a strategy to place this new edge
judiciously (resp.\ adversarially) so that the error exponent deteriorates
as little (resp.\ as much) as possible?

To do so, we say  edge $e$ {\em contains} node $v$ if $e=(v,i)$ and we define the nodes in the smaller tree $T_p$
\begin{align}
v_{\min}^*&:=\argmin_{v\in\calV} \max_{ e \in \calE_p}\{|\rho_e|: e \mbox{ contains node } v\} . \label{eqn:minmax_node}\\
v_{\max}^*&:= \argmax_{v\in\calV} \max_{e \in
\calE_p}\{|\rho_e|: e \mbox{ contains node }
v\} .\label{eqn:maxmax_node}
\end{align}
\begin{proposition}[Error Exponent of Larger Trees] \label{prop:supertree}
Assume that $|\rho_{\mathrm{new}}| < |\rho_e| \, \forall\, e \in \calE_p$. Then,
\begin{itemize}
\item[(a)] The difference between the error exponents $\tilK_p-\tilK_{p''}$ is {\it minimized} when
$T_{p''}$ is obtained by adding to $T_p$ a new edge with correlation
coefficient $\rho_{\mathrm{new}}$ at vertex $v_{\min}^*$ given by~\eqref{eqn:minmax_node} as a leaf.
\item[(b)] The difference $\tilK_p-\tilK_{p''}$ is {\it maximized} when the new edge is added  to $v_{\max}^*$  given by~\eqref{eqn:maxmax_node} as a leaf.
\end{itemize}
\end{proposition}
\begin{IEEEproof} The vertex given by~\eqref{eqn:minmax_node} is the best vertex  to attach the new edge by Lemma~\ref{thm:mono}(b).    
\end{IEEEproof} 
This result implies that if we receive data dimensions sequentially, we have a straightforward rule in~\eqref{eqn:minmax_node} for identifying larger trees such that the exponent decreases as little as possible at each step. 


\section{Numerical Experiments}\label{sec:num}

We now perform experiments with the following two objectives.
Firstly, we study the accuracy of the Euclidean approximations
(Theorem~\ref{thm:euc_gauss}) to identify regimes in which
the approximate crossover rate $\tilJ_{e,e'}$ is close to the true
crossover rate $J_{e,e'}$. Secondly, by performing simulations we
study how various tree structures ({\it e.g.} chains and stars)
influence the error exponents (Theorem~\ref{thm:graph_topo}).

\subsection{Comparison Between True and Approximate Rates} \label{sec:compare_true_approx}
\begin{figure}
\centering
\includegraphics[width=2.8in]{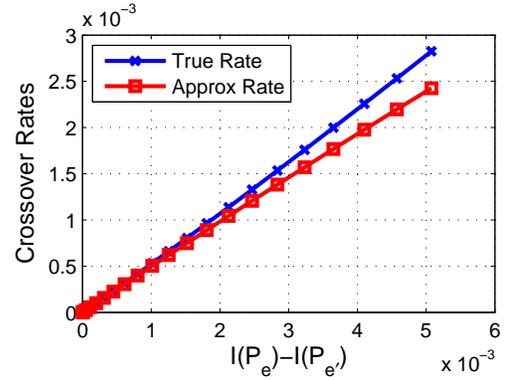}
\caption{Comparison of true and approximate  crossover rates in~\eqref{eqn:ldp_gauss}  and~\eqref{eqn:Jee_gauss} respectively.  }
\label{fig:trueapprox}
\end{figure}

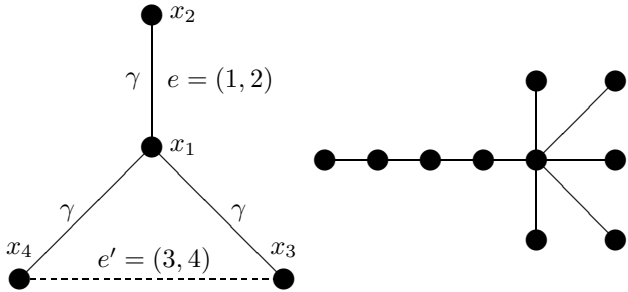
\begin{figure}
\centering
\begin{tabular}{lr}
\begin{picture}(100,100)
\put(0,0){\line(1,1){50}}
\put(100,0){\line(-1,1){50}}
\put(50,50){\line(0,1){50}}
\put(0,0){\circle*{8}}
\put(50,50){\circle*{8}}	
\put(100,0){\circle*{8}}
\put(50,100){\circle*{8}}
\put(0,11){\makebox (0,0){$x_4$}}
\put(62,50){\makebox (0,0){$x_1$}}
\put(100,11){\makebox (0,0){$x_3$}}
\put(62,100){\makebox (0,0){$x_2$}}
\put(76,75){\makebox (0,0){$e=(1,2)$}}
\put(51,7){\makebox (0,0){$e'=(3,4)$}}
\put(43,75){\makebox (0,0){$\gamma$}}
\put(18,25){\makebox (0,0){$\gamma$}}
\put(83,25){\makebox (0,0){$\gamma$}}
\multiput(0,0)(4,0){25}{\line(1,0){2}}
\end{picture} &
\begin{picture}(100,100)
\put(0,45){\circle*{8}}
\put(20,45){\circle*{8}}
\put(40,45){\circle*{8}}
\put(60,45){\circle*{8}}
\put(80,45){\circle*{8}}
\put(110,45){\circle*{8}}
\put(110,75){\circle*{8}}
\put(110,15){\circle*{8}}
\put(80,75){\circle*{8}}
\put(80,15){\circle*{8}}
\put(0,45){\line(1,0){110}}
\put(80,15){\line(0,1){60}}
\put(80,45){\line(1,1){30}}
\put(80,45){\line(1,-1){30}}
\end{picture}
\end{tabular}
\caption{Left: The symmetric star graphical model used for comparing the true and approximate crossover rates as described in Section~\ref{sec:compare_true_approx}. Right: The structure of a  {\em hybrid}  tree graph with $d=10$ nodes as described in Section~\ref{sec:compare_diff_trees}. This is a tree with  a length-$d/2$ chain and a order $d/2$ star attached to one of the leaf nodes of the chain.}
\label{fig:star}
\end{figure}

In Fig.~\ref{fig:trueapprox}, we plot the {\em true} and {\em approximate} crossover rates\footnote{This small example has sufficient illustrative power because as we have seen, errors occur locally and only involve triangles.   }  (given in  \eqref{eqn:ldp_gauss} and \eqref{eqn:tilJee} respectively) for a 4-node symmetric star graph, whose structure is shown in Fig.~\ref{fig:star}. The zero-mean Gaussian graphical model has a covariance matrix $\bSigma$ such that $\bSigma^{-1}$ is parameterized by $\gamma\in(0,1/\sqrt{3})$ in the following way: $\bSigma^{-1}(i,i) = 1$ for all $i$, $\bSigma^{-1}(1,j) = \bSigma^{-1}(j,1) = \gamma$ for all $j=2,3,4$ and $\bSigma^{-1}(i,j)=0$ otherwise. By increasing $\gamma$, we increase the difference of the mutual information quantities on the edges $e$ and non-edges $e'$. We see from
Fig.~\ref{fig:trueapprox} that both rates increase as the
difference $I(p_e) - I(p_{e'})$ increases. This is in line with our
intuition because if $p_{e,e'}$ is such that $I(p_e) - I(p_{e'})$
is large, the crossover rate is also  large. We also observe that if
$I(p_e) - I(p_{e'})$ is small, the true and approximate rates are
close. This is also in line with the assumptions of
Theorem~\ref{thm:euc_gauss}. When the difference between the mutual
information quantities increases, the true and approximate rates
separate from each other.

\subsection{Comparison of Error Exponents Between  Trees} \label{sec:compare_diff_trees}
\begin{figure}
  \centering
  \subfigure 
  {
      \includegraphics[width=2.7in]{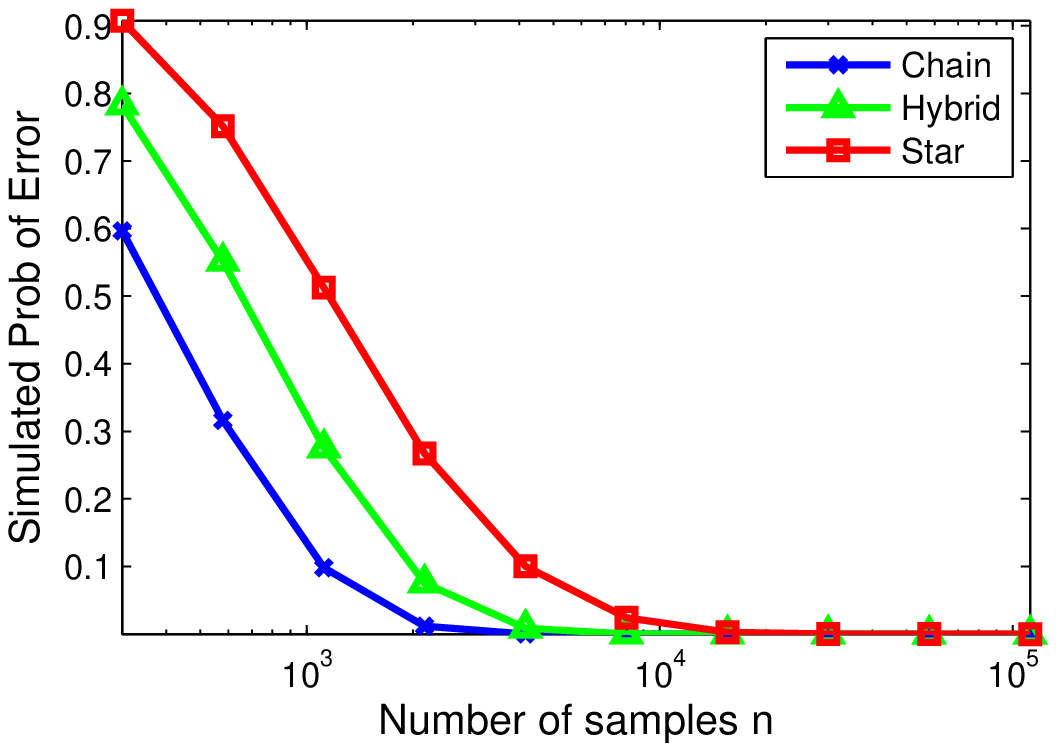}
      \label{fig:sims1}
  }      \hspace{.3in}
  \subfigure 
  {
      \includegraphics[width=2.7in]{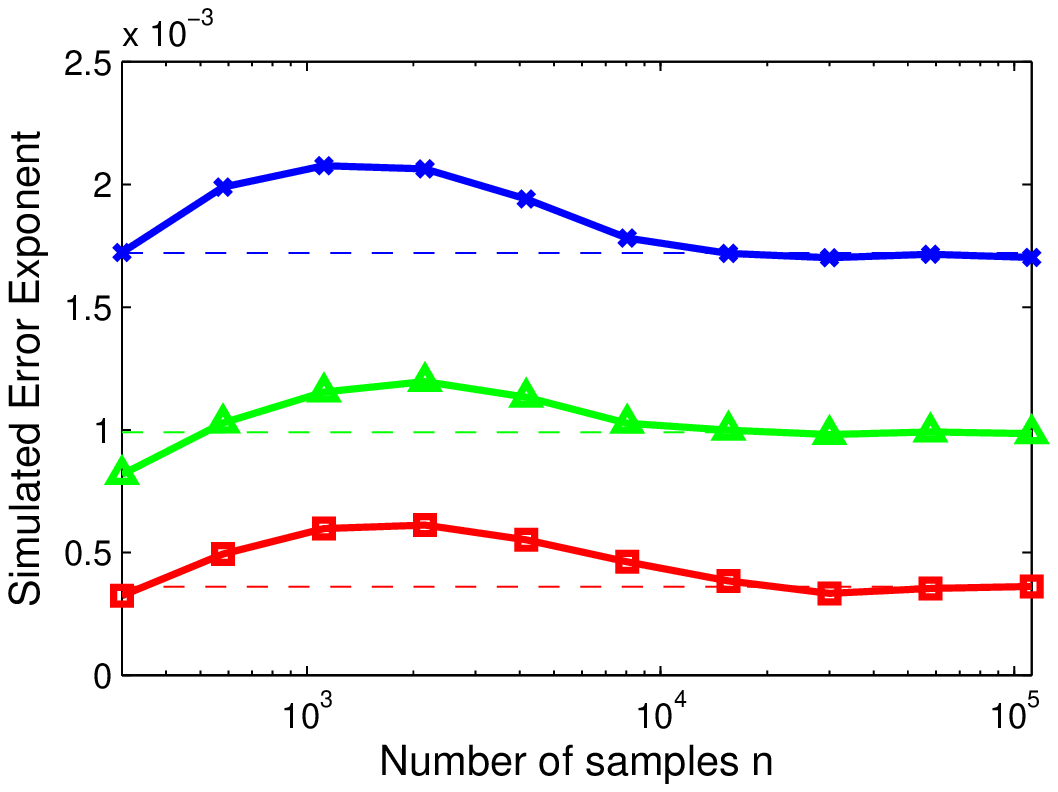}
      \label{fig:sims2}
  }
   \caption{Simulated error probabilities and error exponents for chain, hybrid and star graphs with fixed $\rhobf$. The dashed lines show the true error exponent $K_p$ computed numerically using~\eqref{eqn:ldp_gauss} and~\eqref{eqn:Jfinal2}. Observe that the simulated error exponent converges to the true error exponent as $n\to \infty$. The legend   applies to both plots.}
  \label{fig:sims}
\end{figure}

In Fig.~\ref{fig:sims}, we simulate  error probabilities by drawing i.i.d.\ samples from three  $d=10$ node tree graphs -- a chain, a star and a hybrid between a chain and a star as shown in Fig.~\ref{fig:star}. We then used the samples to learn the structure via the Chow-Liu procedure~\cite{CL68} by solving the MWST in~\eqref{eqn:mwst}. The $d-1=9$  correlation coefficients were chosen to be  equally spaced in the interval $[0.1,0.9]$ and they were randomly placed on the edges of the three tree graphs. We observe from Fig.~\ref{fig:sims} that for fixed $n$, the star and chain have the highest and lowest error probabilities $\bP(\calA_n)$ respectively. The {\it simulated error exponents} given by  $\{-n^{-1} \log \bP(\calA_n)\}_{n\in\bN}$ also converge to their true values as $n\to\infty$. The exponent associated to the star is higher than that of the chain, which is corroborated by  Theorem~\ref{thm:graph_topo}, even though the theorem only applies in the very-noisy case (and for $|\rho_i|<0.63055$ in the case of the chain). From this experiment, the claim also seems to be true even though the setup is not very-noisy. We also observe that the error  exponent of the hybrid  is between that of the star and the chain.

\section{Conclusion}\label{sec:concl}
Using the theory of large deviations, we have obtained the error exponent associated with learning the structure of a Gaussian tree model. Our analysis in this theoretical paper also answers the fundamental questions as to which set of parameters and which structures result in high and low error exponents. We conclude that Markov chains (resp.\ stars) are the easiest (resp.\ hardest) structures to learn as they maximize (resp.\  minimize) the error exponent. Indeed, our numerical experiments on a variety of Gaussian graphical models validate the theory presented. We believe the intuitive results presented in this paper will lend useful insights for modeling high-dimensional data using tree distributions.





\section*{Acknowledgments}
The authors would like to acknowledge Profs.\ Lang Tong, Lizhong Zheng and  Sujay Sanghavi for extensive
discussions. The authors would also like to acknowledge the anonymous reviewer who found an error  in
 the original manuscript that led to the revised development leading to Theorem~\ref{thm:graph_topo}.

\appendices

\section{Proof of Theorem \ref{thm:ldp_gauss_mi}} \label{app:ldp_mi}

\begin{IEEEproof}
This proof   borrows ideas from~\cite{Shen07}. We assume $m=4$ (\ie, disjoint edges) for simplicity. The result for $m=3$ follows similarly.  Let $\calV'\subset \calV$ be a set of $m=4$ nodes corresponding to node pairs $e$ and $e'$. 
Given a subset of node pairs $\mathcal{Y}\subset \calV' \times\calV'$ such that
$(i,i)\in \mathcal{Y}, \forall\, i\in \calV'$, the set of {\em feasible moments}~\cite{Wai08}  is defined as
\begin{align}
\mathcal{M}_{\mathcal{Y}}~:=~ \big\{ & \bmeta_{e,e'} \in \bR^{|\mathcal{Y}|}~:~ \exists \, q (\cdot) \in \calP(\bR^m) \nonumber\\
&\text{    s.t.    }\,\, \bE_q[x_ix_j]=\eta_{i,j}, \forall \, (i,j)\in
\mathcal{Y} \big\}. \label{eqn:marg_poly}
\end{align}
Let the set of densities with moments $\bmeta_{e,e'} :=\{\eta_{i,j}\!:\!(i,j)\in\mathcal{Y} \} 	$ be denoted as
\begin{eqnarray}
\calB_{\mathcal{Y}}(\bmeta_{e,e'})\!:=\! \{q\!\in\!
\calP(\bR^m)\!:\!\bE_q[x_i x_j]\!=\!\eta_{i,j},   (i,j)\in
\mathcal{Y}\}. \label{eqn:BS}
\end{eqnarray}
\begin{lemma}[Sanov's Thm, Contraction Principle~\cite{Deu00}] \label{lem:ldp_cont}
For the event that the empirical moments of the i.i.d.\ observations $\bx^n$ are equal to $\bmeta_{e,e'}=\{\eta_{i,j}:(i,j)\in\mathcal{Y}\}$, we have the LDP
\begin{align}
\lim_{n\rightarrow\infty}-\frac{1}{n}\log \bP& \left[
\bigcap_{(i,j)\in\mathcal{Y}} \Bigg\{\bx^n:\frac{1}{n}\sum_{k=1}^n x_{k,i}x_{k,j} = \eta_{i,j}\Bigg\}
\right]  \nonumber\\
&~=~\inf_{q_{e,e'}\in  \calB_{\mathcal{Y}}(\bmeta)} D(q_{e,e'} \, ||\, p_{e,e'}).
\label{eqn:gauss_gen}
\end{align}
If $\bmeta_{e,e'} \in \mathcal{M}_{\mathcal{Y}}$, the  optimizing pdf
$q_{e,e'}^*$ in~\eqref{eqn:gauss_gen} is given by $  q_{e,e'}^*(\bx)\propto p_{e,e'}(\bx)\exp \big[ \sum_{(i,j)\in \mathcal{Y}} \theta_{i,j}\, x_i x_j \big] ,$  where the set of constants $\{\theta_{i,j}:(i,j)\in\mathcal{Y}\} $ are chosen such that
$q_{e,e'}^* \in \calB_{\mathcal{Y}}(\bmeta_{e,e'})$ given in \eqref{eqn:BS}.
\end{lemma}

From Lemma~\ref{lem:ldp_cont}, we conclude that the optimal $q_{e,e'}^*$ in~\eqref{eqn:gauss_gen} is a Gaussian. Thus, we can restrict our search for the optimal distribution to a search over Gaussians, which are parameterized by means and covariances. The crossover event for mutual information defined in~\eqref{eqn:crossover} is
$
\calC_{e,e'} =\left\{\hrho_{e'}^2\ge \hrho_e^2 \right\},
$
since in the Gaussian case, the mutual information is a monotonic function of the square of the correlation coefficient (cf.\ Eqn.~\eqref{eqn:simple_mi}). Thus it suffices to consider $\left\{\hrho_{e'}^2\ge \hrho_e^2 \right\}$, instead of the event involving the mutual information quantities. Let $e=(i,j)$, $e'=(k,l)$ and $\bmeta_{e,e'} :=(\eta_e, \eta_{e'}, \eta_{i}, \eta_{j} , \eta_{k}, \eta_{l})\in \mathcal{M}_{\mathcal{Y}} \subset \bR^6  $ be the moments of $p_{e,e'}$, where $\eta_e:=\bE[x_i x_j]$ is the covariance of $x_i$ and $x_j$, and $\eta_{i}:=\bE[x_i^2]$ is the variance of  $x_i$ (and similarly for the other moments). Now apply the contraction principle~\cite[Ch.\ 3]{Den00} to the continuous map $h:\mathcal{M}_{\mathcal{Y}}  \to\bR$, given by the difference between the square of correlation coefficients 
\begin{equation}
h(\bmeta_{e,e'}):= \frac{\eta_e^2}{\eta_{i}\eta_{j}  }  - \frac{\eta_{e'}^2}{\eta_{k}\eta_{l} }.  \label{eqn:h_cont}
\end{equation}
Following the same argument as in \cite[Theorem 2]{Tan09}, the equality case dominates $\calC_{e,e'}$, \ie, the event $\left\{\hrho_{e'}^2= \hrho_e^2 \right\}$ dominates  $\left\{\hrho_{e'}^2\ge \hrho_e^2 \right\}$.\footnote{This is also intuitively true because the most likely way the error event $\calC_{e,e'}$ occurs is when equality  holds, \ie, $\left\{\hrho_{e'}^2= \hrho_e^2 \right\}$.} Thus, by considering the set $\{\bm{\eta}_{e,e'}:h(\bm{\eta}_{e,e'})=0\}$, the rate corresponding to  $\calC_{e,e'}$ can be written as
\begin{equation}
J_{e,e'} = \inf_{\bmeta_{e,e'} \in\mathcal{M}_{\mathcal{Y}}}\left\{ g(\bmeta_{e,e'}) :   \frac{\eta_e^2}{ {\eta_{i} \eta_{j}} }  =  \frac{\eta_{e'}^2}{ {\eta_{k}\eta_{l}}}  \right\}, \label{eqn:ldp_gauss1}
\end{equation}
where the function $g:\mathcal{M}_{\mathcal{Y}} \subset \bR^6 \!\to\! [0,\infty)$ is defined as
\begin{equation}
g(\bmeta_{e,e'}):=\inf_{q_{e,e'} \in \calB_{\mathcal{Y}}(\bmeta_{e,e'})}\, D(q_{e,e'} \,||\, p_{e,e'}),\label{eqn:Jbar}
\end{equation}
and the set $\calB_{\mathcal{Y}}(\bmeta_{e,e'})$ is defined in~\eqref{eqn:BS}. Combining expressions in~\eqref{eqn:ldp_gauss1} and~\eqref{eqn:Jbar} and the fact that the optimal solution $q_{e,e'}^*$ is Gaussian yields $J_{e,e'}$ as given in the statement of the theorem (cf.\ Eqn.~\eqref{eqn:ldp_gauss}).

The second assertion in the theorem follows from the fact that since $p_{e,e}$   satisfies $I(p_e)\ne I(p_{e'})$, we have $|\rho_e|\ne |\rho_{e'}|$ since $I(p_e)$ is a monotonic function in $|\rho_e|$. Therefore, $q_{e,e'}^* \ne p_{e,e'}$ on a set whose (Lebesgue) measure $\nu $ is strictly positive. Since $D(q_{e,e'}^*||p_{e,e'})=0$ if and only if $q_{e,e'}^*=p_{e,e'}$ almost everywhere-$[\nu]$, this implies that $D(q_{e,e'}^*||p_{e,e'})>0$~\cite[Theorem 8.6.1]{Cov06}.
\end{IEEEproof}

\section{Proof of Corollary \ref{thm:Jge0_gauss}}\label{app:greaterzero}
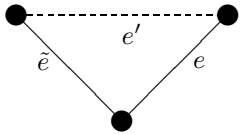
\begin{figure}
\centering
\begin{picture}(80, 40)
\put(0,40){\circle*{8}}
\put(40,0){\circle*{8}}
\put(80,40){\circle*{8}}

\put(0,40){\line(1,-1){40}}
\put(40,00){\line(1,1){40}}
 
\multiput(0,40)(4,0){20}{\line(1,0){2}}

\put(8,19){\mbox{$\tilde{e}$}}
\put(67,20){\mbox{${e}$}}
\put(40,29){\mbox{$e'$}}
\end{picture}
\caption{Illustration for the proof of Corollary~\ref{thm:Jge0_gauss}. The correlation coefficient on the non-edge is $\rho_{e'}$ and satisfies $|\rho_{e'}|=|\rho_e|$ if $|\rho_{\tilde{e}}|=1$.  }
\label{fig:triangle_rho}
\end{figure}

\begin{IEEEproof}
($\Rightarrow$)  Assume that $K_p>0$. Suppose, to the contrary, that either (i) $T_p$ is a forest or (ii) $\mathrm{rank}(\bSigma)<d$ amd $T_p$ is not a forest. In (i), structure estimation of $p$ will be inconsistent (as described in Section~\ref{sec:chowliu}), which implies that $K_p = 0$, a contradiction. In (ii), since $p$ is a spanning tree, there exists an edge $\tilde{e}\in\calE_p$ such that the  correlation coefficient $\rho_{\tilde{e}}= \pm 1$ (otherwise $\bSigma$ would be full rank). In this case, referring to Fig.~\ref{fig:triangle_rho} and assuming that $|\rho_e|\in (0,1)$, the correlation on the non-edge $e'$ satisfies $|\rho_{e'}|=|\rho_e||\rho_{\tilde{e}}|=|\rho_e|$, which implies that $I(p_e)=I(p_{e'})$. Thus, there is no unique maximizer in~\eqref{eqn:mwst} with the empiricals $\hp_e$ replaced by $p_e$.  As a result, ML for structure learning via~\eqref{eqn:mwst} is inconsistent hence $K_p=0$, a contradiction. 
 
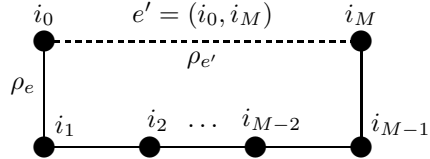
\begin{figure}
\centering
\begin{picture}(130, 50)
\put(0,0){\line(1,0){120}}
\put(0,0){\line(0,1){40}}
\put(120,0){\line(0,1){40}}
\put(0,0){\circle*{8}}
\put(40,0){\circle*{8}}
\put(80,0){\circle*{8}}
\put(120,0){\circle*{8}}
\put(0,40){\circle*{8}}
\put(120,40){\circle*{8}}
\put(0,50){\makebox (0,0){$i_0$}}
\put(120,50){\makebox (0,0){$i_M$}}
\put(135,8){\makebox (0,0){$i_{M-1}$}}
\put(8,8){\makebox (0,0){$i_1$}}
\put(43,10){\makebox (0,0){$i_2$}}
\put(60,8){\makebox (0,0){$\ldots$}}
\put(86,10){\makebox (0,0){$i_{M-2}$}}
\put(60,33){\makebox (0,0){$\rho_{e'}$}}
\put(-8,23){\makebox (0,0){$\rho_{e}$}}
\put(60,50){\makebox (0,0){${e'} =  (i_0,i_{M})$}}
\multiput(0,40)(4,0){30}{\line(1,0){2}}
\end{picture}
\caption{Illustration for the proof of Corollary \ref{thm:Jge0_gauss}. The unique path between $i_0$ and $i_M$ is $(i_0, i_1,\ldots, i_M)=\Path(e';\calE_p)$. }
\label{fig:Kge0}
\end{figure}

($\Leftarrow$) Suppose both $\bSigma\succ 0$ and $T_p$ {\em not} a proper forest, \ie, $T_p$ is a spanning tree. Assume, to the contrary, that $K_p=0$. Then from~\cite{Tan09}, $I(p_e)=I(p_{e'})$ for some $e'\notin\calE_p$ and some $e\in\Path(e';\calE_p)$. This implies that $|\rho_e| =|\rho_{e'}|$. Let $e' =(i_0,i_M)$ be a non-edge and let the unique path from node $i_0$ to node $i_M$ be $(i_0, i_1,\ldots, i_M)$ for some $M\ge 2$. See Fig.~\ref{fig:Kge0}. Then,  
$
|\rho_{e'}| = |\rho_{i_0,i_M}| = |\rho_{i_0,i_1}||\rho_{i_1,i_2}|\ldots |\rho_{i_{M-1},i_M}|. 
$
Suppose, without loss of generality, that edge $e=(i_0,i_1)$ is such that $ |\rho_{e'}|=|\rho_{e}|$ holds, then we can cancel $|\rho_{e'}|$ and $ |\rho_{i_0,i_1}|$ on both sides   to give
$
|\rho_{i_1,i_2}||\rho_{i_2,i_3}| \ldots |\rho_{i_{M-1},i_M}|=1. 
$
Cancelling $\rho_{e'}$ is legitimate because we assumed that $\rho_{e'}\ne 0$ for all $e'\in \calV\times\calV$, because  $p$ is a {\it spanning} tree. Since each correlation coefficient has magnitude not exceeding 1, this means that each correlation coefficient has magnitude 1, \ie, $|\rho_{i_1,i_2}|=\ldots=|\rho_{i_{M-1},i_M}|= 1$. Since the  correlation coefficients equal to $\pm 1$, the  submatrix of the covariance matrix $\Sigma$ containing these correlation coefficients is not positive definite. Therefore by Sylvester's condition, the covariance matrix $\bSigma\nsucc 0$, a contradiction. Hence, $K_p>0$. 
\end{IEEEproof}

\section{Proof of Theorem \ref{thm:euc_gauss}} \label{app:Jee}
\begin{IEEEproof}
We first assume that $e$ and $e'$ do not share a node. The approximation of the KL-divergence for Gaussians can be written as in \eqref{eqn:gauss_app_obj}. We now linearize the constraint set $L_{\bDelta}(p_{e,e'}) $ as defined in \eqref{eqn:constraint_set}. Given a positive definite covariance matrix $\mathbf\bSigma_e\in\bR^{2\times 2}$, to simplify the notation, let $I(\bSigma_e)=I(\calN(\bx;
\bzero,\bSigma_e))$ be the mutual information of the two random variables  with covariance matrix $\bSigma_e$. We now perform a first-order Taylor expansion of the mutual information around  $\bSigma_e$. This can be expressed as
\begin{eqnarray}
I(\bSigma_e+\bDelta_e)\!=\! I(\bSigma_e)\!+\! \Tr\left( \nabla_{\bSigma_e} I(\bSigma_e)^T \bDelta_e \right)\!+\!o(\bDelta_e). \label{eqn:taylor_I}
\end{eqnarray}
Recall that the Taylor expansion of log-det~\cite{Faz03} is
$
\log \det(\mathbf{A}) = \log \det(\mathbf{B})+ \langle\mathbf{A}-\mathbf{B}, \mathbf{B}^{-1}\rangle + o(\| \mathbf{A}-\mathbf{B}\|_F) ,
$
with the notation $\langle\mathbf{A}-\mathbf{B}, \mathbf{B}^{-1}\rangle=\Tr( (\mathbf{A}-\mathbf{B}) \mathbf{B}^{-1})$. Using this result we can conclude that the gradient of $I$ with respect to $\bSigma_e$ in the above expansion \eqref{eqn:taylor_I} can be simplified to give the matrix
\begin{equation}
\nabla_{\bSigma_e} I(\bSigma_e) = -\frac{1}{2} \begin{pmatrix}
0 & [\bSigma_e^{-1}]_{od} \\ [\bSigma_e^{-1}]_{od} & 0 \end{pmatrix},
\end{equation}
where $[\mathbf{A}]_{od}$ is the (unique) off-diagonal element of the $2\times 2$ symmetric matrix $\mathbf{A}$. By applying the same expansion to $I(\bSigma_{e'}+\bDelta_{e'})$, we can express the linearized constraint as
\begin{equation}
\langle \bM, \bDelta\rangle = \Tr(\bM^T  \bDelta)= I(\bSigma_e)-I(\bSigma_{e'}),\label{eqn:gauss_lin_constr}
\end{equation}
where the symmetric matrix $\bM =  \bM(\bSigma_{e,e'})$ is defined in the following fashion: $\bM(i,j) =   \frac{1}{2}[\bSigma_e^{-1}]_{od}$ if $(i,j) = e$, $\bM(i,j) = -\frac{1}{2}[\bSigma_{e'}^{-1}]_{od}$ if $(i,j) = e'$ and $\bM(i,j)=0$ otherwise. 

Thus, the problem reduces to minimizing (over $\bDelta$) the approximate objective in~\eqref{eqn:gauss_app_obj} subject to the linearized constraints in~\eqref{eqn:gauss_lin_constr}. This is a least-squares problem. By using the matrix derivative identities
$
\nabla_{\bDelta}\Tr(\bM \bDelta) = \bM $ and $
\nabla_{\bDelta}\Tr((\bSigma^{-1} \bDelta)^2) = 2\bSigma^{-1}\bDelta \bSigma^{-1},
$
we can solve for the optimizer $\bDelta^*$ yielding:
\begin{equation}
\bDelta^* = \frac{I(\bSigma_e)-I(\bSigma_{e'}) }{(\Tr(\bM \bSigma ) )^2 } \bSigma \bM \bSigma .
\end{equation}
Substituting the expression for $\bDelta^*$ into~\eqref{eqn:gauss_app_obj} yields
\begin{equation}
\widetilde{J}_{e,e'} = \frac{(I(\bSigma_e)-I(\bSigma_{e'}))^2}{4\,\Tr((\bM \bSigma)^2)}=  \frac{(I(p_e)-I(p_{e'}))^2}{4\,\Tr((\bM \bSigma)^2)}.\label{eqn:Jee_tr}
\end{equation}
Comparing \eqref{eqn:Jee_tr} to our desired result \eqref{eqn:Jee_gauss}, we observe that problem now reduces to showing that $\Tr((\bM \bSigma)^2) = \frac{1}{2} \var(s_e-s_{e'})$. To this end, we note that for Gaussians, the information density is
$
s_e(x_i,x_j) = -\frac{1}{2}\log(1-\rho_e^2)-  [\bSigma_e^{-1}]_{od}\,x_i\, x_j .\nn
$
Since the first term is a constant, it  suffices to compute $\var(  [\bSigma_e^{-1}]_{od}x_i x_j -   [\bSigma_{e'}^{-1}]_{od}\,x_k\, x_l  )$.
Now, we define the matrices
\begin{equation}
\bC \!:=\! \begin{pmatrix} 0 & 1/2\\ 1/2 &0 \end{pmatrix}, \,\, \bC_1  \!:=\! \begin{pmatrix}\bC & \bzero\\ \bzero & \bzero  \end{pmatrix}, \,\,\bC_2  \!:=\! \begin{pmatrix}\bzero & \bzero\\ \bzero & \bC  \end{pmatrix},\nn
\end{equation}
and use the following identity for the normal random vector $(x_i,x_j, x_k, x_l)\sim\calN(\bzero, \bSigma)$
\begin{align}
\mathrm{Cov}(a x_i x_j, b x_k x_l) = 2ab \cdot\Tr(\bC_1 \bSigma\bC_2 \bSigma),\quad \forall \, a,b\in\bR,\nn
\end{align}
and the definition of $\bM$ to conclude that $\var(s_e-s_{e'})=2\Tr((\bM\bSigma)^2)$.  This completes the proof for the case when $e$ and $e'$ do not share a node. The proof for the case when $e$ and $e'$ share a node proceeds along exactly the same 
lines with a slight modification of the matrix $\bM$. 
\end{IEEEproof}

\section{Proof of Lemma \ref{thm:mono}} \label{app:mono}
\begin{IEEEproof}
Denoting the correlation coefficient on edge $e$ and non-edge $e'$ as  $\rho_e$ and $\rho_{e'}$ respectively, the approximate crossover rate can be expressed as
\begin{equation}
\tilJ (\rho_e, \rho_{e'}) = \frac{A(\rho_e^2,
\rho_{e'}^2)}{B(\rho_e^2, \rho_{e'}^2)}, \label{eqn:tildeJee}
\end{equation}
where  the numerator and the denominator  are defined as
\begin{align}
A(\rho_e^2,\rho_{e'}^2) &:= \left[\frac{1}{2} \log \left( \frac{1 \!- \!\rho_{e'}^2}{1 \!- \!\rho_{e}^2} \right)\right]^2 , \nn \\
B(\rho_e^2,\rho_{e'}^2) \! :=\! \frac{2(\rho_{e'}^4+\rho_{e'}^2)}{(1-\rho_{e'}^2)^2}  \!&+\! \frac{2(\rho_{e}^4+\rho_{e}^2)}{(1-\rho_{e}^2)^2} \!-\! \frac{4\rho_{e'}^2 (\rho_{e}^2+1)}{ (1-\rho_{e'}^2) (1-\rho_{e}^2)} .\nn
\end{align}
The evenness result follows  from $A$ and $B$   because
$\tilJ (\rho_e, \rho_{e'})$ is, in fact a function of $(\rho_e^2,
\rho_{e'}^2)$. To simplify the notation, we make the following
substitutions: $x:= \rho_{e'}^2$ and $y :=  \rho_e^2$. 
Now we apply the quotient rule to \eqref{eqn:tildeJee}. Defining  $\mathcal{R}:= \{(x,y)\in \bR^2: y\in (0,1), x\in (0,y)\},$ it suffices to show that
$$
C(x,y):= B(x,y) \frac{\partial A(x,y)}{\partial x} -  A(x,y) \frac{\partial B(x,y)}{\partial x}\le 0,
$$
for all $(x,y)\in \mathcal{R}$. Upon simplification, we have
$$
C(x,y) \!=\!  \frac{  \log \left(\frac{1-x}{1-y}\right) \left[ \log \left(\frac{1-x}{1-y}\right) C_1(x,y) \!+\! C_2(x,y) \right] }{2(1-y)^2(1-x)^3},
$$
where $C_1(x,y) \!:=\! y^2x-6xy-1-2y+3y^2$ and $C_2(x,y) \!:=\! 2x^2 y -6x^2+2x-2y^2 x+8xy -2y-2y^2$. Since $x < y$, the logs in $C(x,y)$ are positive, \ie, $  \log \left(\frac{1-x}{1-y}\right)>0$, so it suffices to show that
$$
\log \left(\frac{1-x}{1-y}\right) C_1(x,y) + C_2(x,y) \le 0.
$$
for all $(x,y)\in \mathcal{R}$. By using the inequality $\log(1+t)\le t$ for all $t > -1$, it again suffices to show that
$$
C_3(x,y):= (y-x)C_1(x,y)+ (1-y)C_2(x,y)\le 0.
$$
Now upon simplification, $C_3(x,y)= 3y^3x-19y^2 x -3y-2y^2+5y^3  -3y^2x^2+14x^2y + 3x+8xy-6x^2 ,$
and this polynomial is equal to zero in  $\overline{\mathcal{R}}$
(the closure of $\mathcal{R}$) iff  $x = y$. At all other points in $\mathcal{R}$, $C_3(x,y)<
0$. Thus, the derivative of $\tilJ(\rho_e, \rho_{e'})$ with respect to
 $\rho_{e'}$ is indeed strictly negative on $\mathcal{R}$. Keeping
$\rho_e$ fixed, the function $\tilJ (\rho_e, \rho_{e'})$ is
monotonically decreasing in $\rho_{e'}^2$ and hence $|\rho_{e'}|$.
Statements (c) and (d) follow along exactly the same lines and are omitted
for brevity.
\end{IEEEproof}

\section{Proofs of Theorem \ref{thm:graph_topo} and Corollary~\ref{cor:best_dist}}\label{app:bestworst}

\begin{IEEEproof}
Proof of $T_{p_{\min}(\rhobf)} = T_{\mathrm{star}}(d)$:  Sort the correlation coefficients in decreasing order of magnitude and relabel the edges such that $|\rho_{e_1}| \ge  \ldots \ge  | \rho_{e_{d-1}}|.$  Then, from  Lemma~\ref{thm:mono}(b), the set of crossover rates for the star graph is given by
$
\{\tilJ (\rho_{e_1}, \rho_{e_1}\rho_{e_2})\} \cup
\{\tilJ(\rho_{e_i}, \rho_{e_i}\rho_{e_1}): i=2, \ldots, d-1\}.$
For edge $e_1$, the   correlation coefficient $\rho_{e_2}$ is the largest correlation coefficient (and hence results in the smallest rate). For all other edges $\{e_i:i\ge 2\}$, the  correlation coefficient $\rho_{e_1}$ is the largest possible correlation coefficient (and hence results in the smallest rate). Since each member in the set of crossovers is the minimum possible, the minimum of these crossover rates is also the minimum possible among all tree graphs.
\end{IEEEproof}

%
Before we prove part (b), we present some properties of the edge weights  $W (\rho_i, \rho_j)$, defined in \eqref{eqn:wts}. 
\begin{lemma}[Properties of Edge Weights] \label{lem:edge_weights}
Assume that all the correlation coefficients are bounded above by $ \rho_{\mathrm{crit}}$, \ie,  $|\rho_i|\le \rho_{\mathrm{crit}}$. Then $W(\rho_i, \rho_j) $  satisfies the following properties:
\begin{enumerate}
\item[(a)] The weights are symmetric, \ie, $W(\rho_i, \rho_j) = W(\rho_j, \rho_i) $.
\item[(b)]  $W(\rho_i, \rho_j)  = \tilJ( \min\{|\rho_i|,| \rho_j|\},\rho_i\rho_j)$, where $\tilJ$ is the approximate crossover rate given in~\eqref{eqn:tildeJee}.
\item[(c)] If $|\rho_i|\ge  |\rho_j|\ge |\rho_k|$, then 
\begin{equation} 
W(\rho_i,\rho_k) \le \min\{ W(\rho_i, \rho_j) , W(\rho_j, \rho_k)\}. \label{eqn:wts_ineq1}
\end{equation}
\item[(d)] If $|\rho_1|\ge \ldots \ge  |\rho_{d-1}|$, then
\begin{subequations} \label{eqn:wts_ineq}
\begin{align}
W(\rho_i,\rho_j)&\le W(\rho_i,\rho_{i+1}) ,\quad\forall\, j\ge i+1, \label{eqn:wts_ineq2}\\
W(\rho_i,\rho_j)&\le W(\rho_i,\rho_{i-1}) ,\quad\forall\, j\le i-1. \label{eqn:wts_ineq3}
\end{align}
\end{subequations}
\end{enumerate}
\end{lemma}
\begin{IEEEproof}
Claim (a) follows directly from the definition of $\tilJ$  in~\eqref{eqn:wts}. Claim  (b) also follows from the definition of $\tilJ$ and its monotonicity property in Lemma~\ref{thm:mono}(d). Claim  (c) follows by first using Claim (b) to establish  that the RHS of~\eqref{eqn:wts_ineq1} equals $\min\{\tilJ(\rho_j,\rho_j\rho_i), \tilJ(\rho_k,\rho_k\rho_j)\}$ since $|\rho_i|\ge  |\rho_j|\ge |\rho_k|$. By the same argument, the LHS of~\eqref{eqn:wts_ineq1},  equals  $\tilJ(\rho_k,\rho_k\rho_i)$. Now we have
\begin{eqnarray}
\tilJ(\rho_k,\rho_k\rho_i)\le \tilJ(\rho_j,\rho_j \rho_i),  \quad
\tilJ(\rho_k,\rho_k\rho_i)\le \tilJ(\rho_k,\rho_k \rho_j),  \label{eqn:useLemma5}
\end{eqnarray}
where the first and second inequalities follow from Lemmas~\ref{thm:mono}(c) and~\ref{thm:mono}(b) respectively. This establishes~\eqref{eqn:wts_ineq1}. Claim  (d) follows by  applying Claim  (c) recursively. 
\end{IEEEproof}
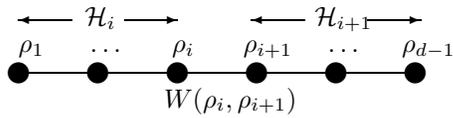
\begin{figure}
\centering
\begin{picture}(150, 35)
\put(0,10){\circle*{8}}
\put(30,10){\circle*{8}}
\put(60,10){\circle*{8}}
\put(90,10){\circle*{8}}
\put(120,10){\circle*{8}}
\put(150,10){\circle*{8}}
\put(0,10){\line(1,0){150}}
\put(0,18){\mbox{$\rho_1$}}
\put(58,18){\mbox{$\rho_i$}}
\put(85,18){\mbox{$\rho_{i+1}$}}
\put(145,18){\mbox{$\rho_{d-1}$}}

\put(27,18){\mbox{$\ldots$}}
\put(117,18){\mbox{$\ldots$}}

\put(55,-3){\mbox{$W(\rho_i,\rho_{i+1})$}}

\put(20,30){\vector(-1,0){20}}
\put(40,30){\vector(1,0){20}}

\put(108,30){\vector(-1,0){20}}
\put(132,30){\vector(1,0){20}}

\put(25,28){\mbox{$\calH_i$}}
\put(112,28){\mbox{$\calH_{i+1}$}}

\end{picture}
\caption{Illustration of the proof of Theorem~\ref{thm:graph_topo}. Let $|\rho_1|\ge \ldots \ge |\rho_{d-1}|$. The figure shows the chain $H_{\mathrm{chain}}^*$ (in the line graph domain) where the correlation coefficients $\{\rho_i\}$ are placed in decreasing order.  }
\label{fig:cyclic}
\end{figure}

\begin{IEEEproof}
Proof of $T_{p_{\max}(\rhobf)} = T_{\mathrm{chain}}(d)$:  Assume, without loss of generality, that $|\rho_{e_1}|\ge  \ldots \ge  |\rho_{e_{d-1}}|$ and we also abbreviate $\rho_{e_i}$ as $\rho_i$ for all $i=1,\ldots, d-1$. We use the idea of line graphs introduced in Section~\ref{sec:basics_graphs} and Lemma~\ref{lem:edge_weights}. Recall that $\calL(\calT^d)$ is the set of line graphs of spanning trees with $d$ nodes.  From~\eqref{eqn:pmax_line},  the line graph for the structure of the best distribution $p_{\max,\rhobf}$ for learning in~\eqref{eqn:max_prob} is
\begin{equation}
H_{\max,\rhobf} ~:=~ \argmax_{H \in \calL(\calT^d)}  \min_{  (i,j)\in H} W(\rho_i,\rho_j). \label{eqn:hmax_line}
\end{equation}
We now argue that the length $d-1$ chain $H_{\mathrm{chain}}^*$ (in the line graph domain) with correlation coefficients  $\{\rho_i\}_{i=1}^{d-1}$ arranged in decreasing order on the nodes (see Fig.~\ref{fig:cyclic}) is the line graph that optimizes~\eqref{eqn:hmax_line}.  Note that the edge weights of $H_{\mathrm{chain}}^*$ are given by $W(\rho_i,\rho_{i+1})$ for $ 1\le i\le d-2$. Consider any other line graph $H\in\calL(\calT^d)$. Then we claim that 
\begin{eqnarray}
\min_{(i,j)\in H\setminus H_{\mathrm{chain}}^*} W(\rho_i,\rho_j) \le \min_{(i,j)\in H_{\mathrm{chain}}^*\setminus H } W(\rho_i,\rho_j). \label{eqn:Hsetminus}
\end{eqnarray}
To prove~\eqref{eqn:Hsetminus},  note that any edge $(i,j)\in H_{\mathrm{chain}}^*\setminus H$ is {\em consecutive}, \ie, of the form $(i,i+1)$. Fix any such  $(i,i+1)$.  Define the two subchains of $H_{\mathrm{chain}}^*$ as $\calH_{i}:=\{(1,2),\ldots, (i-1,i)\}$ and $\calH_{i+1}:=\{(i+1,i+2),\ldots, (d-2,d-1)\}$ (see Fig.~\ref{fig:cyclic}). Also, let $\calV(\calH_{i}):=\{1,\ldots, i\}$ and $\calV(\calH_{i+1}):=\{i+1,\ldots, d-1\}$ be the nodes in subchains  $\calH_{i}$ and $\calH_{i+1}$ respectively.  Because $(i,i+1)\notin H$, there is a set of edges (called cut set edges) $\calS_{i}:=\{(j,k)\in H: j\in \calV(\calH_{i}), k\in \calV(\calH_{i+1})\}$  to ensure that the line graph $H$ remains connected.\footnote{The line graph $H=\calL(G)$ of a connected graph $G$ is connected. In addition, any $H\in\calL(\calT^d)$ must be a claw-free, block graph~\cite[Theorem 8.5]{Harary72}.} The  edge weight of each cut set edge $(j,k)\in\calS_i$  satisfies  $W(\rho_j,\rho_k)\le  W(\rho_i,\rho_{i+1})$ by~\eqref{eqn:wts_ineq} because $|j-k|\ge 2$ and $j\le i$ and $k\ge i+1$. By considering all cut set edges $(j,k)\in \calS_{i}$ for fixed $i$ and subsequently all $(i,i+1)\in H_{\mathrm{chain}}^*\setminus H$, we establish~\eqref{eqn:Hsetminus}. It follows that
\begin{equation}
\min_{  (i,j)\in H} W(\rho_i,\rho_j)\le \min_{  (i,j)\in H_{\mathrm{chain}}^*} W(\rho_i,\rho_j),
\end{equation}
because the other edges in $H$ and $H_{\mathrm{chain}}^*$ in \eqref{eqn:Hsetminus} are common.  See Fig.~\ref{fig:example_H} for an example to illustrate~\eqref{eqn:Hsetminus}. 

Since the chain line graph $H_{\mathrm{chain}}^*$ achieves the maximum bottleneck edge weight, it is the optimal line graph, \ie, $H_{\max,\rhobf}=H_{\mathrm{chain}}^*$. Furthermore, since the line graph of a chain is a chain, the best structure $T_{p_{\max}(\rhobf)}$ is also a chain and we have established~\eqref{eqn:bestresult}.  The best distribution is given by the chain with the correlations placed in decreasing order, establishing Corollary~\ref{cor:best_dist}. \end{IEEEproof}
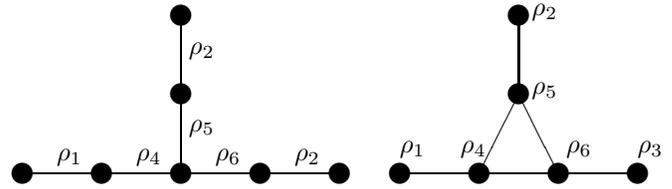
\begin{figure}\centering
\begin{tabular}{cc}
\begin{picture}(120, 60)
\put(0,0){\circle*{8}}
\put(30,0){\circle*{8}}
\put(60,0){\circle*{8}}
\put(90,0){\circle*{8}}
\put(120,0){\circle*{8}}
\put(60,30){\circle*{8}}
\put(60,60){\circle*{8}}
\put(0,0){\line(1,0){120}}
\put(60,0){\line(0,1){60}}
\put(13,5){\mbox{$\rho_{1}$}}
\put(43,5){\mbox{$\rho_{4}$}}
\put(73,5){\mbox{$\rho_{6}$}}
\put(103,5){\mbox{$\rho_{2}$}}
\put(63,16){\mbox{$\rho_{5}$}}
\put(63,45){\mbox{$\rho_{2}$}}
\end{picture} & \hspace{.1in}
\begin{picture}(90, 60)
\put(0,0){\circle*{8}}
\put(30,0){\circle*{8}}
\put(60,0){\circle*{8}}
\put(90,0){\circle*{8}}
\put(45,30){\circle*{8}}
\put(45,60){\circle*{8}}
\put(0,0){\line(1,0){90}}
\put(45,30){\line(0,1){30}}
\put(30,00){\line(1,2){15}}
\put(60,00){\line(-1,2){15}}
\put(0,8){\mbox{$\rho_{1}$}}
\put(23,8){\mbox{$\rho_{4}$}}
\put(63,8){\mbox{$\rho_{6}$}}
\put(90,8){\mbox{$\rho_{3}$}}
\put(50,30){\mbox{$\rho_{5}$}}
\put(50,60){\mbox{$\rho_{2}$}}
\end{picture}
\end{tabular}
\caption{A 7-node tree $T$ and its line graph $H=\calL(T)$ are shown in the left and right figures respectively. In this case $H\setminus H_{\mathrm{chain}}^* =\{(1,4), (2,5), (4,6), (3,6) \}$ and $H_{\mathrm{chain}}^*\setminus H = \{ (1,2), (2,3), (3,4)\}$. Eqn.~\eqref{eqn:Hsetminus} holds because  from~\eqref{eqn:wts_ineq}, $W(\rho_1,\rho_4)\le W(\rho_1,\rho_2)$, $W(\rho_2,\rho_5)\le W(\rho_2,\rho_3)$ etc.\ and also if $a_i\le b_i$ for $i\in\mathcal{I}$ (for finite $\mathcal{I}$), then $\min_{i\in\mathcal{I}}a_i \le \min_{i\in\mathcal{I}}b_i$.}
\label{fig:example_H}
\end{figure}

\bibliographystyle{IEEEbib}
\bibliography{animabib_v2,isitbib}

\end{document}